\documentclass[runningheads]{llncs}
\usepackage[T1]{fontenc}
\usepackage{graphicx}
\usepackage{hyperref}
\usepackage{color}

\usepackage{adjustbox}
\usepackage{multirow}
\usepackage{booktabs} 
\usepackage{makecell}
\usepackage{listings}
\usepackage{subcaption}
\usepackage{todonotes}
\usepackage[inline]{enumitem}
\usepackage[misc,geometry]{ifsym}
\usepackage{rotating}
\usepackage{versions}
\usepackage{url}

\includeversion{extended}
\excludeversion{short}
\def \shortVersionURL{https://doi.org/10.1007/978-3-031-70893-0_21}

\lstdefinelanguage{yaml}{
  basicstyle=\ttfamily\scriptsize,
  breaklines=true,
  keywords={True, False}, 
  keywordstyle=\color{blue}\bfseries\ttfamily\scriptsize,
  moredelim=[is][commentstyle]{||}{££}, 
  identifierstyle=\color{purple}\ttfamily\scriptsize,
  sensitive=false,
  comment=[l]{\#},
  commentstyle=\color{olive}\ttfamily\scriptsize,
  stringstyle=\color{blue}\ttfamily\scriptsize,
  morestring=[b]',
  morestring=[b]",
  literate={0}{{{\color{blue}0}}}1
           {1}{{{\color{blue}1}}}1
           {2}{{{\color{blue}2}}}1
           {3}{{{\color{blue}3}}}1
           {4}{{{\color{blue}4}}}1
           {5}{{{\color{blue}5}}}1
           {6}{{{\color{blue}6}}}1
           {7}{{{\color{blue}7}}}1
           {8}{{{\color{blue}8}}}1
           {9}{{{\color{blue}9}}}1
  }

\begin{document}
\title{Early Explorations of Lightweight Models for Wound Segmentation on Mobile Devices}
\titlerunning{Lightweight Models for Wound Segmentation on Mobile Devices}

\author{Vanessa Borst(\Letter) % \orcidID{0009-0004-7123-7934}
\and Timo Dittus \and % \orcidID{0009-0008-0704-1856} 
Konstantin Müller\and % \orcidID{0000-0001-6540-3124} 
Samuel Kounev } % \orcidID{0000-0001-9742-2063}
\authorrunning{V. Borst et al.}
\institute{
Software Engineering Group, University of Würzburg, Germany\\
\Letter \hspace{0.2cm}Corresponding author: \email{vanessa.borst@uni-wuerzburg.de}}
\maketitle              % typeset the header of the contribution

\begin{extended}\vspace{-0.4cm}\end{extended}
\begin{short}\vspace{-0.4cm}\end{short}

\begin{abstract}
%The abstract should briefly summarize the contents of the paper in 150--250 words.

The aging population poses numerous challenges to healthcare, including the increase in chronic wounds in the elderly. The current approach to wound assessment by therapists based on photographic documentation is subjective, highlighting the need for computer-aided wound recognition from smartphone photos. This offers objective and convenient therapy monitoring, while being accessible to patients from their home at any time. However, despite research in mobile image segmentation, there is a lack of focus on mobile wound segmentation. To address this gap, we conduct initial research on three lightweight architectures to investigate their suitability for smartphone-based wound segmentation. 
%Utilizing public datasets and UNet as a baseline, our results are promising, with the TopFormer and UNeXt variants showing better performance compared to ENet. 
Using public datasets and UNet as a baseline, our results are promising, with both ENet and TopFormer, as well as the larger \mbox{UNeXt} variant, showing comparable performance to UNet. % the more parameter-rich UNet.
Furthermore, we deploy the models into a smartphone app for visual assessment of live segmentation, where results demonstrate the effectiveness of \mbox{TopFormer} in distinguishing wounds from wound-coloured objects. While our study highlights the potential of transformer models for mobile wound segmentation, future work should aim to further improve the mask contours\begin{extended}\footnote{This is an extended version of our KI 2024 paper, which is available at \href{\shortVersionURL}{Springer}.}\end{extended}\begin{short}\footnote{An extended version of this paper is available for download from \href{\extendedVersionURL}{arXiv}.}\end{short}.

\keywords{Medical Image Segmentation \and Mobile AI \and Deep Learning.}
\end{abstract}

\section{Introduction}
\label{sec:introduction}

Chronic wounds, such as diabetic foot ulcers, affect a significant proportion of the world's population and present a major challenge to healthcare systems worldwide. In the United States alone, it is estimated that more than 6 million people suffer from chronic wounds each year~\cite{powers2016wound}. In addition to diabetes, age is a significant risk factor, with a disproportionate impact on the elderly. This is evidenced by a prevalence of 7.8\% in German nursing home residents~\cite{raeder2020prevalence} compared to 1\% in the German population~\cite{heyer2016epidemiology}. As the world's population continues to age and the incidence of conditions such as diabetes increases, the prevalence of chronic wounds is expected to soar in the coming years. This increase poses a significant challenge to healthcare providers in providing appropriate care to those affected.
From the patient's perspective, the need for frequent visits to specialised wound care facilities to manage chronic wounds can be a significant burden. Access to such clinics may be limited, particularly in remote areas, and mobility issues exacerbate the challenge. There is an urgent need for innovative solutions to alleviate these burdens and improve access to wound care services.

Telemedicine and automation offer promising ways to address these challenges by improving efficiency and accessibility. Automated wound size monitoring has the potential to eliminate the need for patients to visit clinics solely for monitoring purposes, thereby reducing the associated logistical challenges and relieving the burden on healthcare professionals. Furthermore, automated approaches provide timely and accurate information on wound status, facilitating proactive intervention.
Mobile semantic segmentation emerges as a key technology in realizing these advancements, providing real-time feedback and segmentation masks directly to users, thereby circumventing latency issues associated with server-based processing. In addition, deploying models directly on smartphones ensures enhanced privacy and security by keeping sensitive medical data local. In addition, it enables offline functionality, which is crucial for applications in remote areas with limited network coverage.
Despite the advantages of mobile segmentation, there are also significant challenges associated with this endeavour. To be effective, such solutions must include small, lightweight models that can be seamlessly deployed on smartphones. At the same time, such models must be robust in the face of challenging conditions such as different backgrounds, varying lighting conditions, and different camera settings, as mobile wound monitoring is intended to take place in the patient's home environment.

However, despite the necessity for compact yet resilient models, the investigation into mobile wound segmentation remains limited to date.
% Related Work on mobile wound segmentation: 
% Erst allgemein Wundsegmentierung, dann mobile Wundsegmentierung (ohne Real-World Deployment) und dann die wenigen mit Deployment
While several methods have been proposed for wound segmentation utilizing feature-engineering-based machine learning~\cite{song2012automated,wang2016area,kolesnik2005multi}, deep learning~\cite{oota2023wsnet,liu2017framework,chino2020segmenting,goyal2017fully}, or even both~\cite{wagh2020semantic,li2018composite}, they often lack emphasis on lightweight architectures and mobile suitability. 
Some publications suggest that the analyzed architecture(s) could be used on mobile devices due to its efficiency, but such indications often rely solely on metrics such as parameter count as mobile performance indicator, lacking real-world deployment~\cite{wang2020fully,oota2023wsnet}.
Of the few approaches deployed to a mobile device in practice, many apply traditional image processing rather than deep learning~\cite{rocha2021woundarch,varma2016vision,ferreira2021experimental}, while others evaluate only single architectures~\cite{ong2020efficient}, or require manual annotation outlines from users~\cite{cazzolato2020URule,cazzolato2021utrack}.
%in the case of commercial apps, lack details of the technical implementation. 
To the best of our knowledge, a systematic comparison of different neural network-based wound segmentation techniques on mobile devices is still pending.
%
% ########################### UPDATED VERSION ##################################
%Conversely, the number of proposed architectures for mobile device segmentation beyond the wound context has increased significantly. This work therefore conducts initial investigations into the applicability of different architectures from non-wound domains for mobile wound segmentation, including a visual comparison in real-world deployment. 
In contrast, the number of proposed architectures for mobile segmentation beyond the wound context has increased significantly. To address this discrepancy, this work conducts initial investigations into the applicability of different architectures from non-wound domains for mobile wound segmentation, including a visual comparison in real-world usage.

\section{Approach}
\label{sec:approach}
\subsection{Task Definition and Requirements}
This study aims to assess the efficacy of lightweight models for mobile wound segmentation by systematically comparing approaches from beyond the wound domain. In particular, we focus on their ability to perform live segmentation from smartphone camera streams, for which we consider the following requirements paramount:
%This study aims to evaluate the efficacy of lightweight models for mobile wound segmentation by systematically comparing approaches from beyond the wound domain, including their ability to perform live segmentation of smartphone camera streams. Real-time mobile wound segmentation primarily necessitates the following requirements:
\begin{enumerate*}[label=(\roman*)]
\item \textbf{accuracy}: High segmentation performance for accurate analysis of wound area %and progression
\item \textbf{efficiency}: Small number of parameters to enable mobile processing
\item \textbf{robustness}: Effective performance under varying conditions for use in non-standardized home environments
\item \textbf{practicality}: Effective segmentation in real-world deployment to enhance medical practice
% Good segmentation performance in real-world deployment to add value to daily medical practice/to create practical added value 
\end{enumerate*}.

\subsection{Model Selection}
The number of proposed models for segmentation on mobile devices has increased significantly beyond wound segmentation. Notable advances include CNN-based methods, such as encoder-decoder architectures with lightweight backbones such as MobileNets~\cite{howard2018mobileNetV2,howard2019mobileNetV3} combined with different decoders, ENet~\cite{paszke2016enet}, UNeXt~\cite{valanarasu2022unext}, or Fast-SCNN~\cite{pudel2019fastscnn}, which have proven successful in mobile environments. In addition, vision transformer-based models, such as MobileFormer~\cite{chen2022mobile}, MobileViT~\cite{mehta2021mobilevit}, Seaformer~\cite{wan2023seaformer}, or TopFormer~\cite{zhang2022topformer}, have been explored increasingly.
From this spectrum, our model selection aims to cover different network types (CNN, vision transformer) and application domains (general-purpose, (bio)medicine), while setting reasonable limits on training effort. Therefore, we limit the inclusion of architectures to four, prioritizing the assessment of different-sized variants of selected methods where feasible.
Specifically, we selected TopFormer~\cite{zhang2022topformer}, a representative (hybrid) vision transformer among recent general-purpose state-of-the-art (SOTA) methods, and UNeXt~\cite{valanarasu2022unext}, a compact CNN-based SOTA technique for medical segmentation. The selection of these two SOTA models was based on the following criteria: %, given the lack of extensive studies on robustness or practicality in the wound context:
\begin{enumerate*}[label=(\roman*)]
\item competitive performance on respective benchmark datasets
%\item Architectural novelty
\item publication in prestigious journals or conferences
\item efficient resource utilization, as indicated by reported parameter counts and GFLOPs
\item code availability on GitHub
\end{enumerate*}.
Additionally, ENet~\cite{paszke2016enet}, a widely referenced general-purpose CNN for real-time semantic segmentation from 2016, was included due to its remarkable efficiency in terms of trainable parameters and promising outcomes in medical segmentation studies~\cite{yuan2023effective,song2024combining}, given its size.
Lastly, to establish a baseline for wound segmentation, we integrated UNet~\cite{ronneberger2015unet}, a prominent model in biomedical segmentation. Table~\ref{tab:approach:selected_methods} outlines the selection, with parameters reported in million (M) for binary segmentation (one out-channel).
%Additionally, we included ENet~\cite{paszke2016enet}, an often-cited general-purpose CNN for real-time semantic segmentation from 2016, for its remarkable efficiency in terms of trainable parameters and its promising results in medical segmentation studies~\cite{yuan2023effective,song2024combining}, given its size. Lastly, to establish a baseline for wound segmentation, we integrated the larger-sized UNet~\cite{ronneberger2015unet}, a highly influential biomedical segmentation model. Table~\ref{tab:approach:selected_methods} provides a summary of the selected methods, with parameters in \mbox{million(M)} for binary segmentation.
%\footnote{Parameters are expressed in \mbox{million(M)} for binary segmentation (one out-channel)}.

\begin{table*}[ht!]
        \centering
	\caption[]{Overview of the included segmentation techniques.}
 %Parameters in \mbox{million(M)} are calculated for binary segmentation (i.e., one out-channel).}
	\label{tab:approach:selected_methods}
        \begin{adjustbox}{max width=\textwidth}
            \begin{tabular}{l@{\hskip 0.2in}c@{\hskip 0.2in}c@{\hskip 0.2in}r@{\hskip 0.1in}r@{\hskip 0.2in}r@{\hskip 0.1in}r@{\hskip 0.1in}l}
                \toprule
                       & \multirow{2}{*}{UNet~\cite{ronneberger2015unet}} & \multicolumn{1}{c}{\multirow{2}{*}{ENet~\cite{paszke2016enet}}} & \multicolumn{2}{c}{UNeXt~\cite{valanarasu2022unext}}     & \multicolumn{3}{c}{TopFormer~\cite{zhang2022topformer}}          \\
                       \cmidrule(lr){4-5}\cmidrule(lr){6-8}
                       &                       & \multicolumn{1}{c}{}                      & Base          & Small         & Base        & Small       & Tiny       \\
                \midrule\midrule
                \textbf{Params} & 31.03M                & 0.35M                                     & 1.47M          & 0.25M          & 5.03M        & 3.01M        & 1.39M       \\
                %\textbf{Year}   & 2015                  & 2016                                      & \multicolumn{2}{c}{2022}      & \multicolumn{3}{c}{2022}               \\
                \textbf{Type}   & CNN                   & CNN                                       & \multicolumn{2}{c}{CNN}       & \multicolumn{3}{c}{Hybrid CNN-ViT} \\
                \textbf{Domain} & Biomedicine             & General purpose                            & \multicolumn{2}{c}{Medicine}  & \multicolumn{3}{c}{General purpose}      \\ 
                \bottomrule
            \end{tabular}
	\end{adjustbox}
\end{table*}

\section{Experimental Setup}
\label{sec:experimental_setup}

\subsection{Dataset}

We perform experiments on a combination of two prominent datasets: the \textit{Foot Ulcer Segmentation Challenge 2021 (FuSeg)}~\cite{wang2024fuseg} and the \textit{Diabetic Foot Ulcer Challenge 2022 (DFUC 2022)}~\cite{kendrick2022translating,yap2024diabetic} datasets. FuSeg encompasses 1,210 foot ulcer images, each with a resolution of 512 × 512 pixels, sourced from 889 patients' medical visits spanning from October 2019 to April 2021. %at the Advancing the Zenith of Healthcare (AZH) Wound and Vascular Center. 
Meanwhile, the DFUC 2022 dataset comprises 4,000 foot ulcer images with a resolution of 640 × 480 pixels, captured approximately 30 to 40 cm away from the ulcer. %, collated through collaboration with multiple hospitals in the United Kingdom and one in New Zealand.
Our custom dataset,~\textit{Combined Foot Ulcers (CFU)}, integrates FuSeg and DFUC 2022, comprising images with mixed resolutions of either 512 × 512 or 640 × 480 pixels. %, standardized to 512 × 512 pixels through padding and resizing during training. 
Due to the privacy of test set annotations (200 out of 1,210 for FuSeg and 2,000 out of 4,000 for DFUC 2022), only publicly accessible training and validation images are utilized for CFU. %our combined dataset.  
To prevent data leakage, duplicate images within FuSeg and DFUC 2022 were identified and removed. A similarity analysis, employing the perceptual hashing algorithm~\cite{micikevicius2018mixed}, was conducted. Image sets with identical raw bytes or perceptual hash values exhibiting a Hamming distance of 11 or less were deemed duplicates. 
\begin{short}For illustrative purposes, a number of examples are provided in the appendix of the \href{\extendedVersionURL}{extended version} of this paper. \end{short}
\begin{extended}For illustrative purposes, a number of examples are provided in Figure~\ref{fig:appendix:duplicates}. \end{extended}
Eliminating duplicate image-annotation pairs, we obtained a final dataset with 2,887 samples after discarding 123 pairs. Next, these samples were randomly partitioned into training (60\%), validation (20\%), and test (20\%) set.

\subsection{Evaluation Protocol}
To assess model performance, we employed established segmentation metrics, such as Dice score (DSC), Intersection over Union (IoU), precision (Prc), and recall (Rec), utilizing the micro-averaging method while excluding the background class. Our analysis covered two training modalities: training models from scratch and utilizing pre-training on either the ImageNet~\cite{deng2009imagenet} or Cityscapes~\cite{cordts2016cityscapes} dataset. To facilitate this, we downloaded pre-trained weights where available: ENet (Cityscapes), UNet (ImageNet), and TopFormer (ImageNet). 
%. Specifically, pre-trained checkpoints were obtained for ENet (Cityscapes), UNet (ImageNet), and TopFormer (ImageNet).  
Since pre-trained weights were unavailable for UNeXt, we trained both variants ourselves for 100 epochs on Cityscapes, using the best checkpoints as initial weights.

\subsection{Implementation Details}
We implemented this approach using Python 3.10.9 and PyTorch 2.0.1 on a server with two partitioned Nvidia A100 GPUs, reserving a large-sized instance with ~40 GB VRAM for our experiments. All models were trained end-to-end using the AdamW optimizer with an initial learning rate of 0.0001, a batch size of four, and the binary cross entropy (BCE) loss. We used a \textit{ReduceLROnPlateau} scheduler conditioned on the Dice score, with a minimum learning rate of 0.000001. All models were trained for 200 epochs from scratch and for at least 130 epochs in the pre-training setting. For both, the best checkpoints, according to the mean IoU on the validation set, were used for the ensuing test set evaluation.
All images were resized to 512 × 512 pixels and normalized during training. 
We applied thresholding to annotated masks to eliminate out-of-class pixels and used the following data augmentations for each training set sample: Gaussian blur with a kernel size of 25 and a randomly sampled standard deviation from a uniform distribution of [0.001, 2.0], random affine transforms (incl. translations by up to 12.5\%, rotations by up to 180 degrees, scaling between 50\% and 150\% of the original size, and shear of up to 22.5 degrees), color jitter, as well as random horizontal and vertical flips, each with a 50\% probability. 
\begin{short}Further details regarding the augmentations are given in the \href{\extendedVersionURL}{extended version} of this paper, using a sample configuration file. \end{short}
\begin{extended}Details about the augmentations can be found in the example configuration file of Listing~\ref{lst:appendix:enet}. \end{extended}

%Images were resized to 512 × 512 pixels and normalized during training. We applied thresholding to annotated masks to eliminate out-of-class pixels and used the following data augmentations: Gaussian blur with a kernel size of 25 and a random standard deviation from [0.001, 2.0], random affine transforms (translations up to 12.5%, rotations up to 180 degrees, scaling between 50% and 150%, shear up to 22.5 degrees), color jitter, and random horizontal and vertical flips, each with a 50% probability. Details about the augmentations can be found in the example configuration file of Listing~\ref{lst:appendix}.
\section{Evaluation}
\label{sec:evaluation}

\subsection{Results}
\label{ssec:eval:results}
We compare the selected models in Table~\ref{tab:eval:comparison_performance}, where we report the micro-averaged metric scores of each variant for both training modalities. % alongside the number of parameters. 
%Parameter amount was calculated for binary semantic segmentation, i.e. one output channel. % at the end of the network. %, and for input images with $512\times512$ resolution.
%
Our findings reveal that, without pretraining, UNeXt-B demonstrates the most favorable segmentation performance in terms of DSC and IoU scores, effectively balancing parameter count and segmentation quality. Notably, all lightweight models, except UNeXt-S, achieve results within a similar size range as the larger-sized UNet, with the base variants of UNeXt and TopFormer, and ENet even slightly surpassing UNet. Pretraining considerably enhances performance across all models regarding IoU and DSC metrics. Specifically, all lightweight networks show comparable performance, with IoU scores ranging from 69\% to 72\% and DSC values from 82\% to 84\%, except for UNeXt-S, which performs worse.

\begin{table*}[ht!]
        \centering
	\caption[]{Comparison of all models on the CFU dataset. The micro-averaged metrics (in \%) were calculated on the test set. Parameters are reported in millions.}
	\label{tab:eval:comparison_performance}
        \begin{adjustbox}{max width=0.95\textwidth}
		\begin{tabular}{l@{\hskip 0.15in}
                            l@{\hskip 0.05in}l@{\hskip 0.05in}l@{\hskip 0.05in}l@{\hskip 0.15in}
                            l@{\hskip 0.05in}l@{\hskip 0.05in}l@{\hskip 0.05in}l@{\hskip 0.15in}
                            l@{\hskip 0.05in}}
			\toprule
                
			&\multicolumn{4}{c}{Without pretraining} 
                &\multicolumn{4}{c}{With pretraining} 
			&  \\
			\cmidrule(lr){2-5}\cmidrule(lr){6-9}
			      Model          & IoU             & DSC              & Prc           & Rec           & IoU              & DSC            & Prc            & Rec         & Params$\downarrow$    \\
			\midrule\midrule            
                U-Net           & 65.69            & 79.30          & 81.57          & \textbf{77.14}         & \textbf{74.69}            & \textbf{85.56}          & 87.74          & 83.39       & 31.03M        \\
                % TopFormer-B     & 66.67            & 80.00          & 84.77          & 75.74         & 72.41            & 84.00          & 84.56          & \textbf{83.44}       & 5.03M          \\
                % TopFormer-B Retrain
                TopFormer-B     & 66.10            & 79.59          & 86.63          & 73.61         & 72.41            & 84.00          & 84.56          & \textbf{83.44}       & 5.03M          \\
                TopFormer-S     & 63.36            & 77.57          & 83.31          & 72.57         & 71.56            & 83.42          & 85.95          & 81.04       & 3.01M         \\
                UNeXt-B         & \textbf{68.96}            & \textbf{81.63}         & 88.49          & 75.76         & 70.12            & 82.43          & 87.40          & 78.01       & 1.47M    \\
                TopFormer-T     & 62.51            & 76.93          & 84.27          & 70.77         & 69.62            & 82.09          & 83.90          & 80.36       & 1.39M           \\
                % ENet Pretrained 200 epochs
                ENet            & 67.27            & 80.43          & \textbf{89.21}          & 72.23         & 71.71            & 83.52          & \textbf{89.98}          & 77.93       & 0.35M    \\
                % ENet Pretrained 130 epochs
                % ENet     Neu  & 67.27            & 80.43          & \textbf{89.21}          & 72.23         & 72.21           & 83.86          & \textbf{87.77}          & 80.28       & 0.35M    \\
                UNeXt-S         & 59.16            & 74.34          & 77.52          & 71.41         & 64.13            & 78.14          & 87.52          & 70.58       & \textbf{0.25M}        \\

                \bottomrule
		\end{tabular}
	\end{adjustbox}
\end{table*}

\subsection{Real-World Deployment}
\label{ssec:eval:real_world_deployment}
\begin{figure}[htb]
    \centering
    \begin{turn}{-90}
        \begin{adjustbox}{max height=\textwidth, max width=\textheight, keepaspectratio}
            \begin{subfigure}{0.165\textwidth}
                \centering
                \includegraphics[angle=90,trim=200 200 200 200, clip, width=0.9\linewidth]{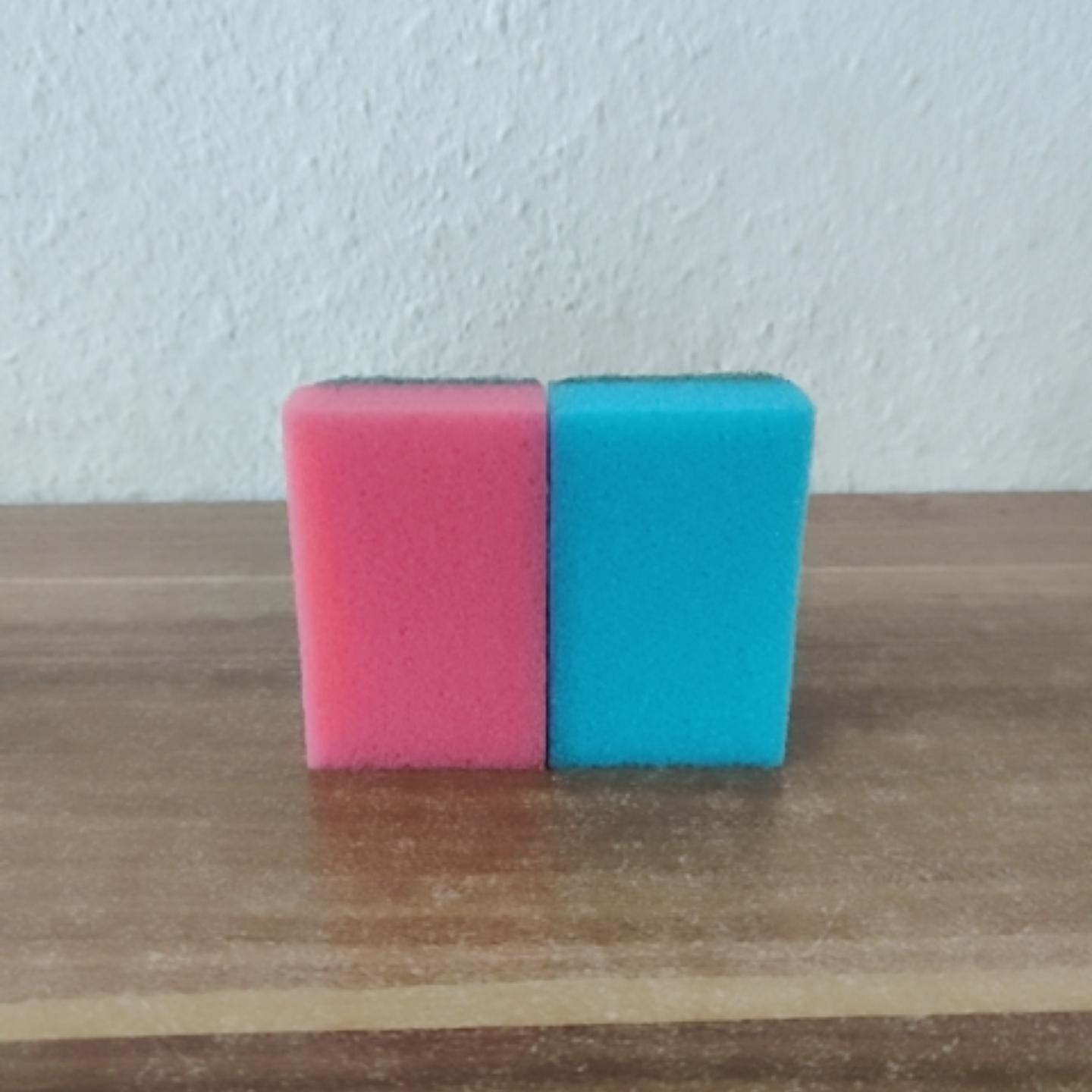}\\[4pt]
                \includegraphics[angle=90,trim=200 200 200 200, clip, width=0.9\linewidth]{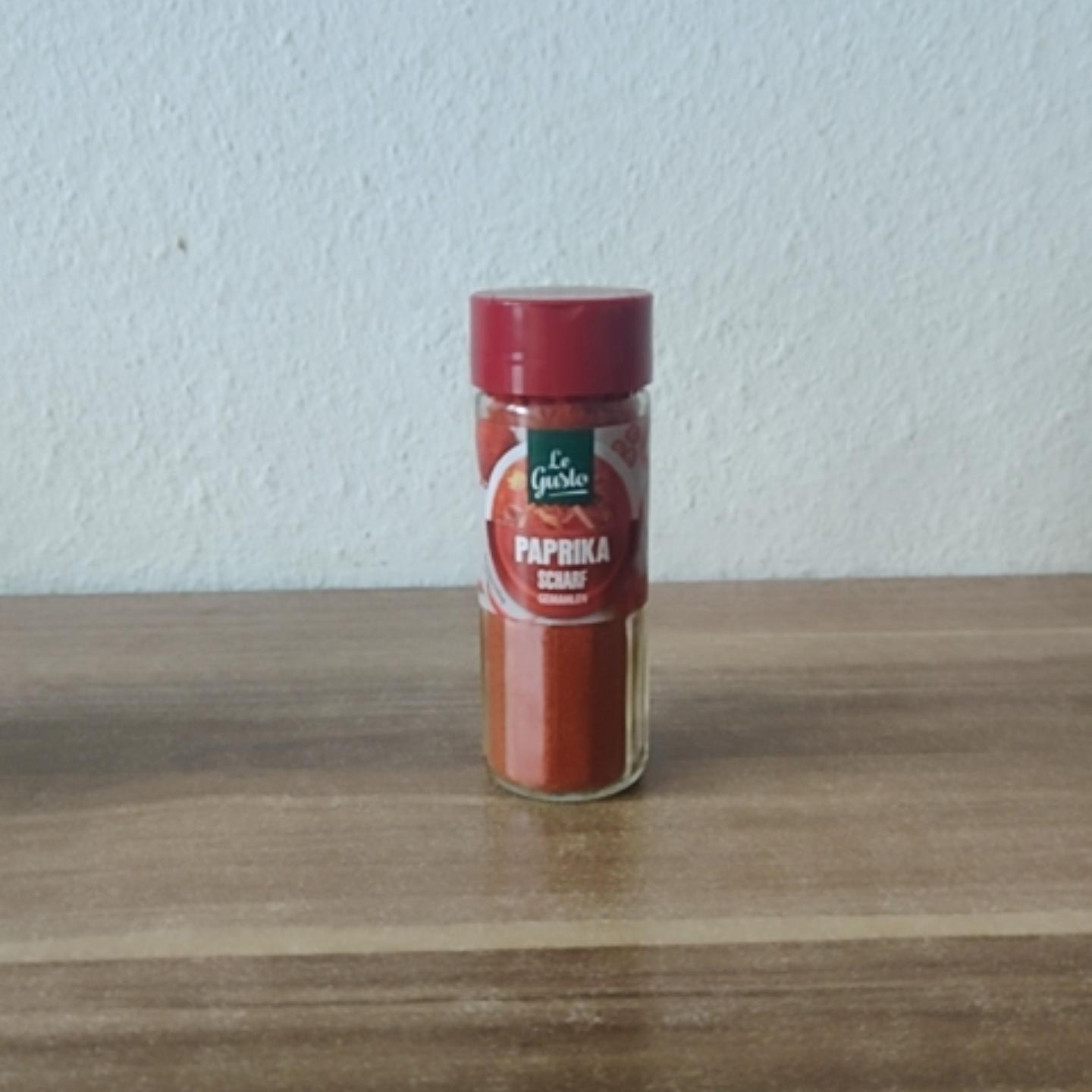}\\[4pt]
                \includegraphics[angle=90,trim=200 200 200 200, clip, width=0.9\linewidth]{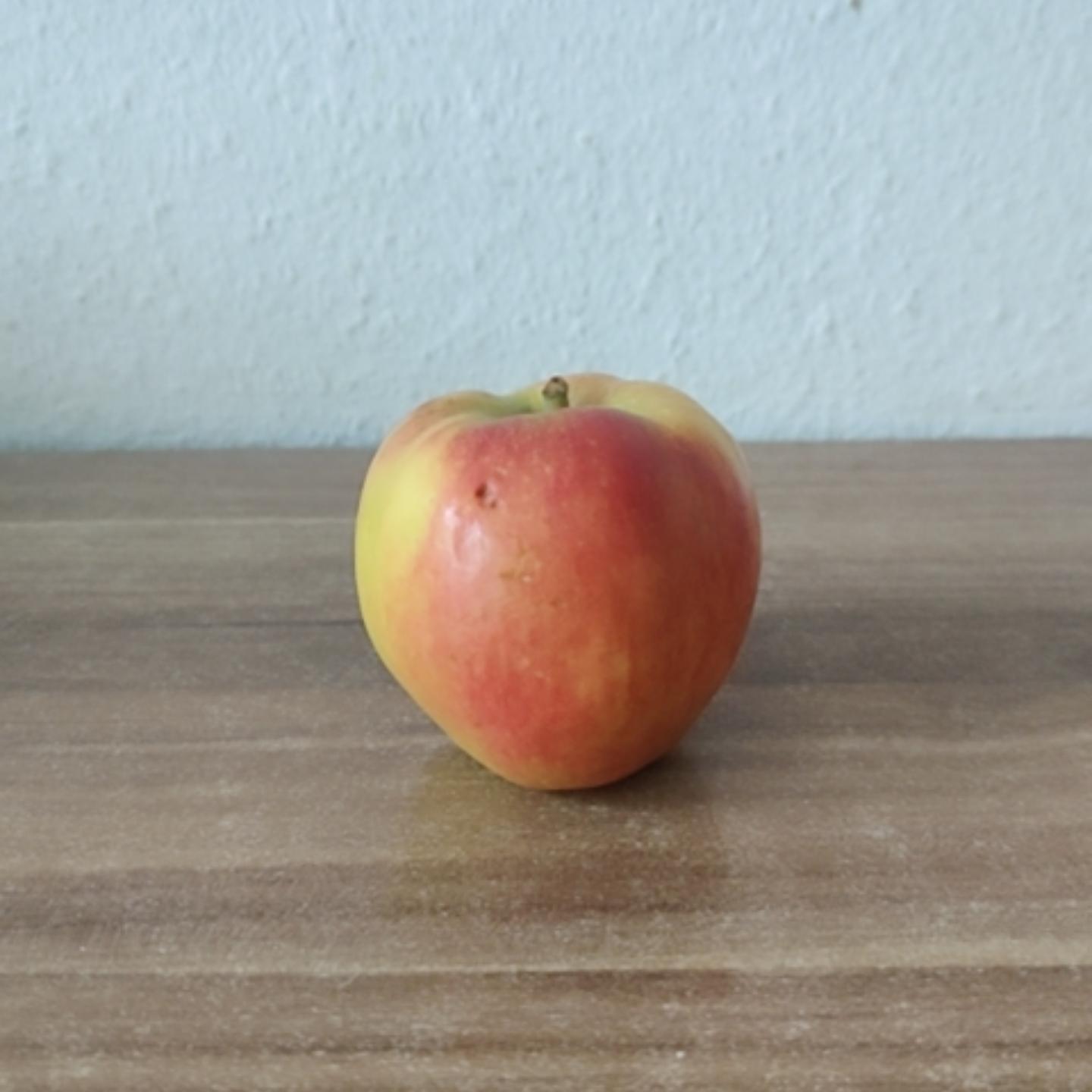}\\[4pt]
                \includegraphics[angle=90,trim=200 200 200 200, clip, width=0.9\linewidth]{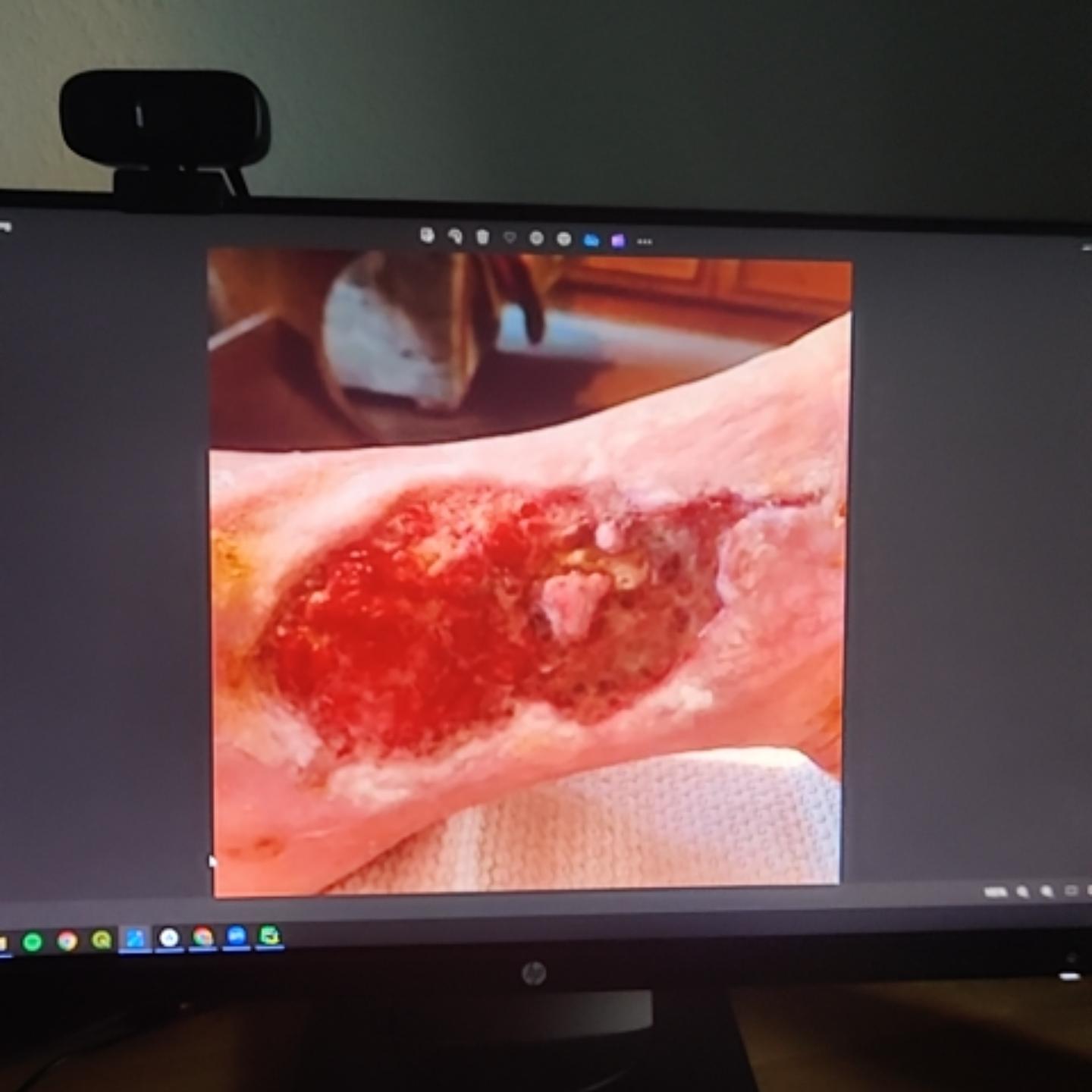}\\[4pt]
                \includegraphics[angle=90,trim=200 200 200 200, clip, width=0.9\linewidth]{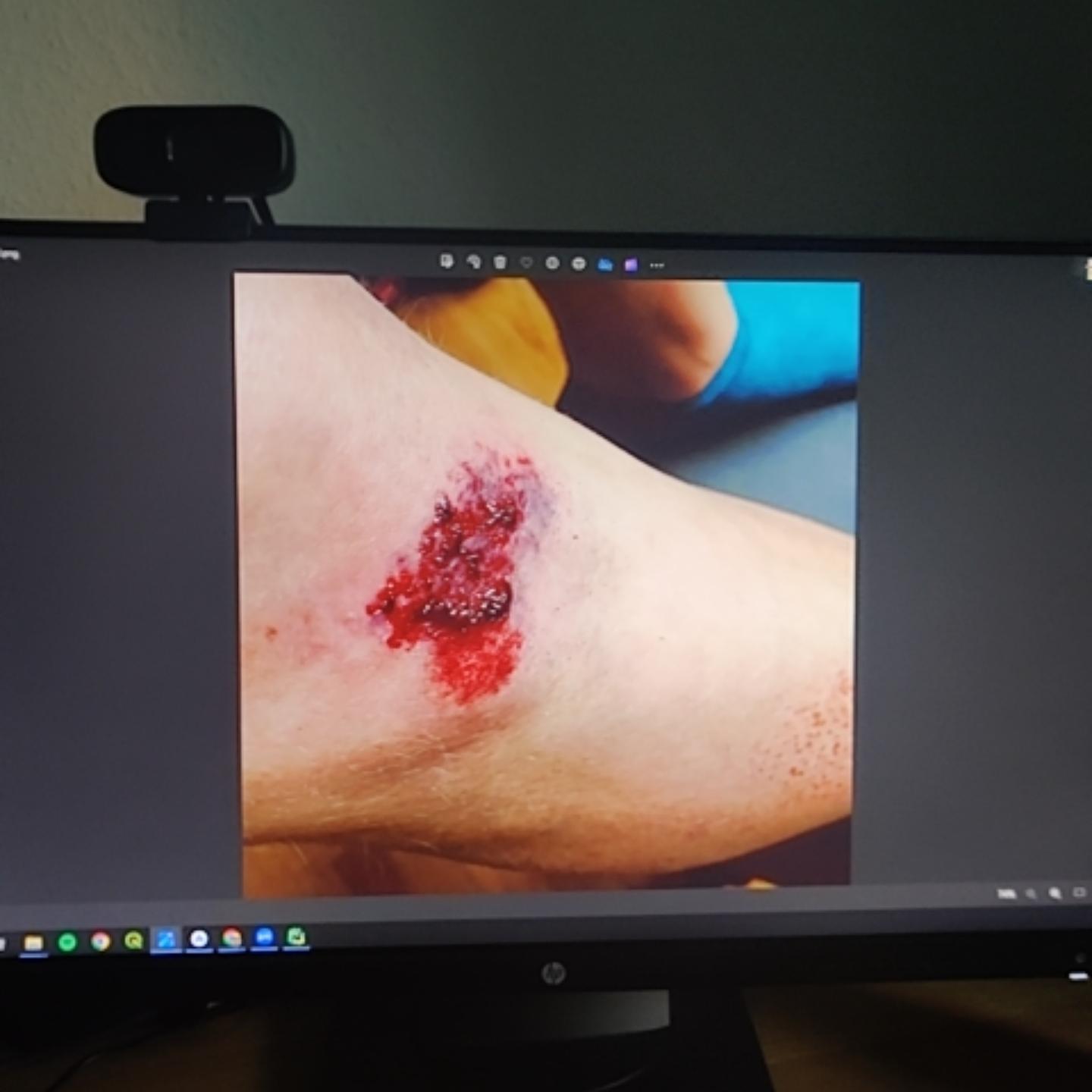}\\[4pt]
                \includegraphics[angle=90,trim=200 200 200 200, clip, width=0.9\linewidth]{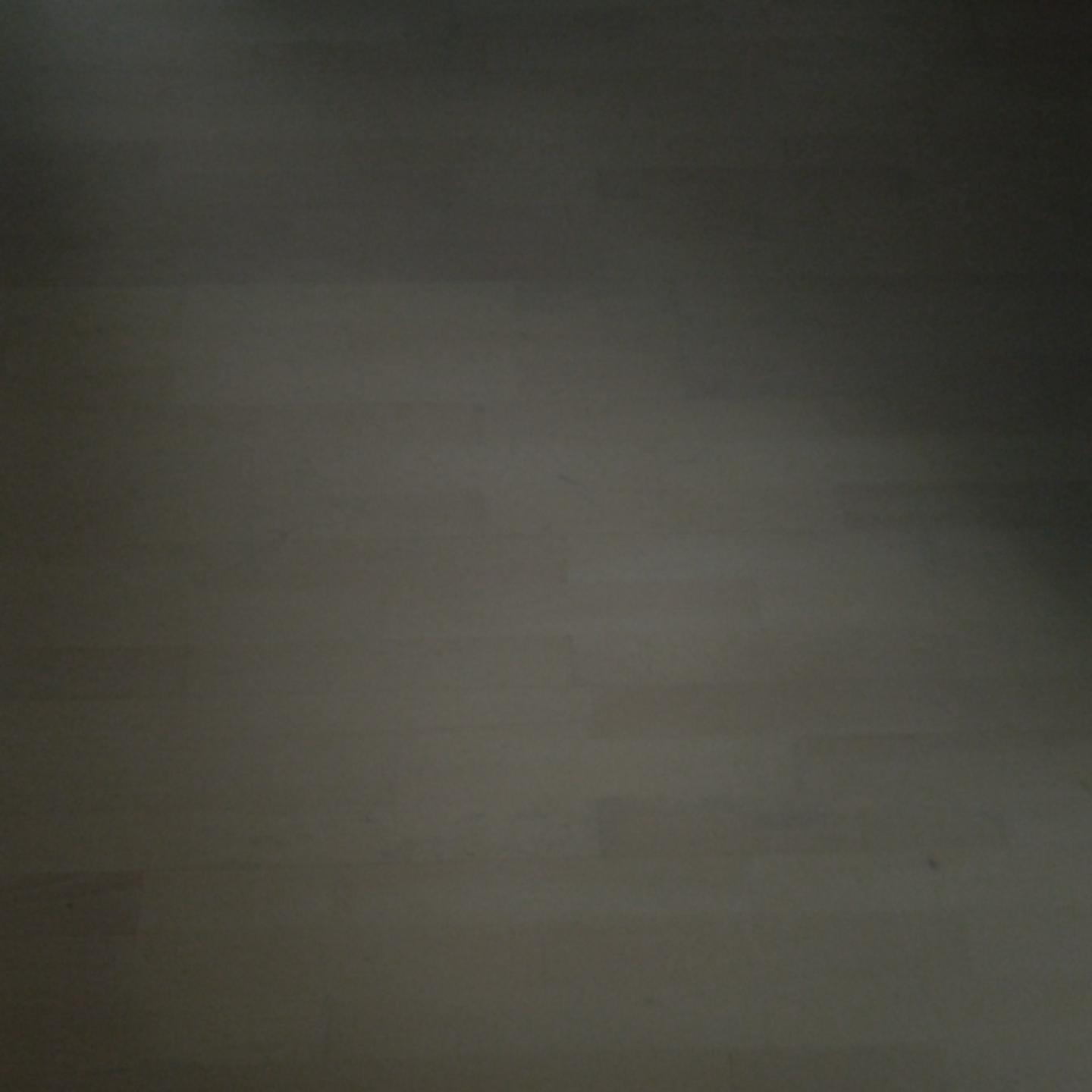}\\[4pt]
                \caption*{\scriptsize Original}
            \end{subfigure}
            \begin{subfigure}{0.165\textwidth}
                \centering
                \includegraphics[angle=90,trim=200 200 200 200, clip, width=0.9\linewidth]{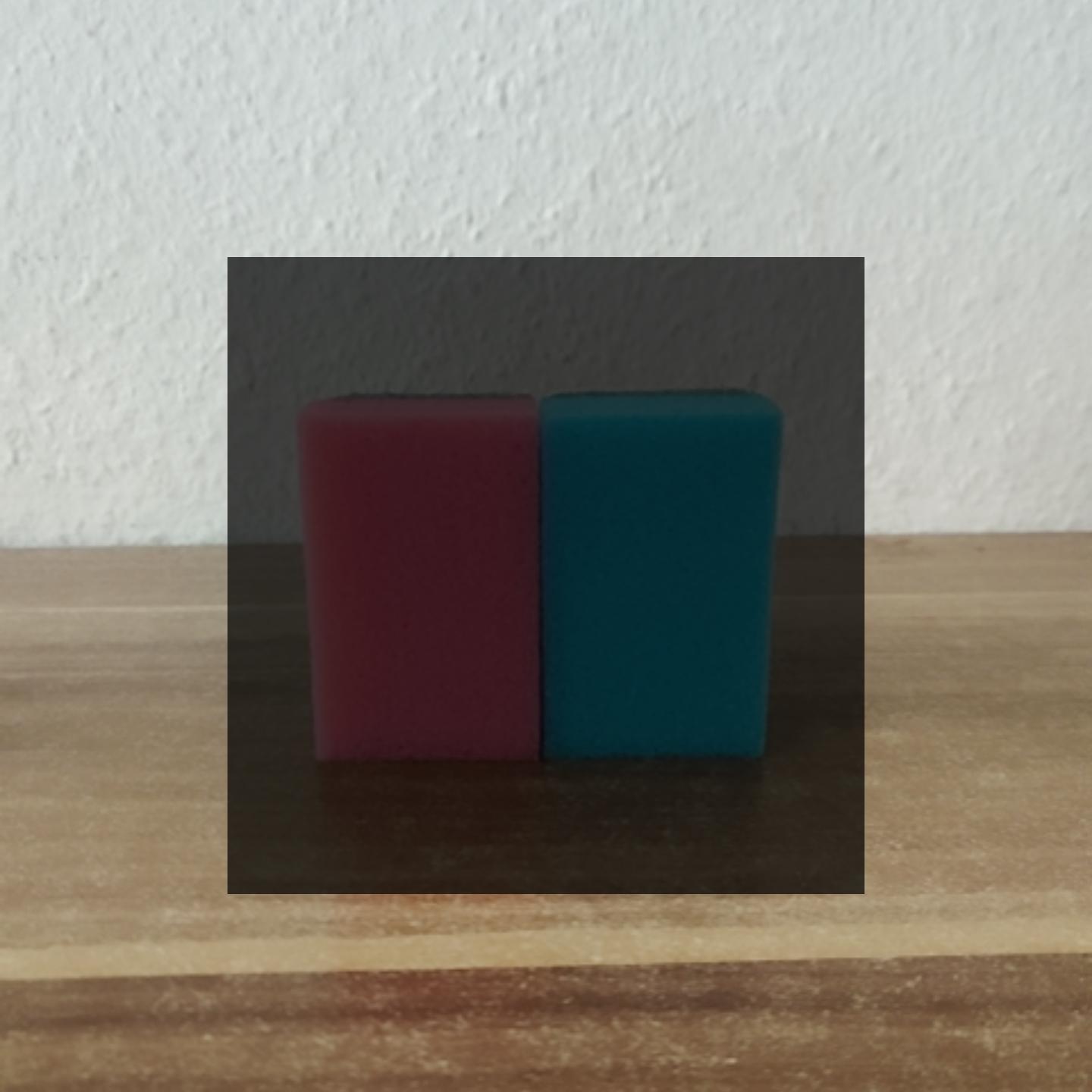}\\[4pt]
                \includegraphics[angle=90,trim=200 200 200 200, clip, width=0.9\linewidth]{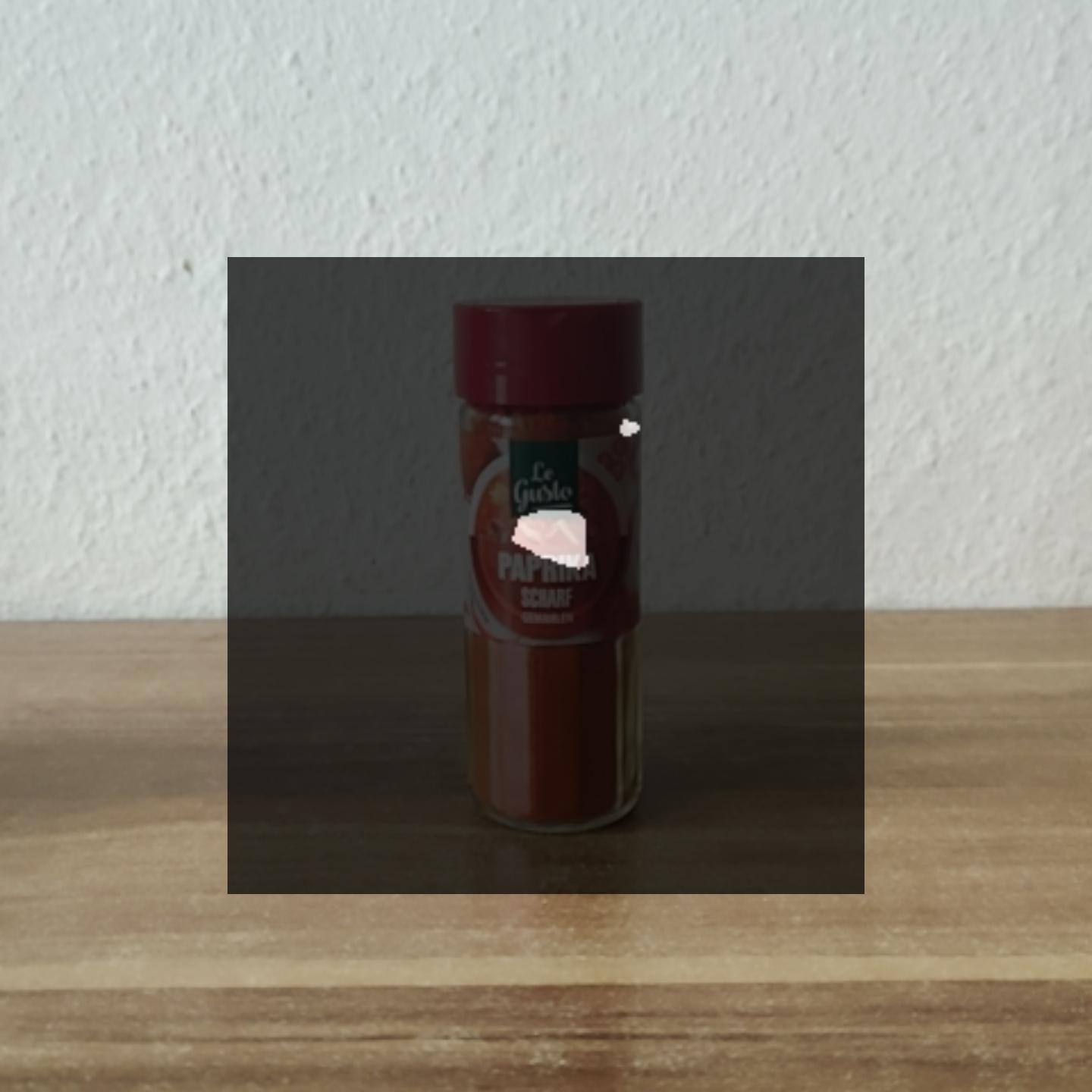}\\[4pt]
                \includegraphics[angle=90,trim=200 200 200 200, clip, width=0.9\linewidth]{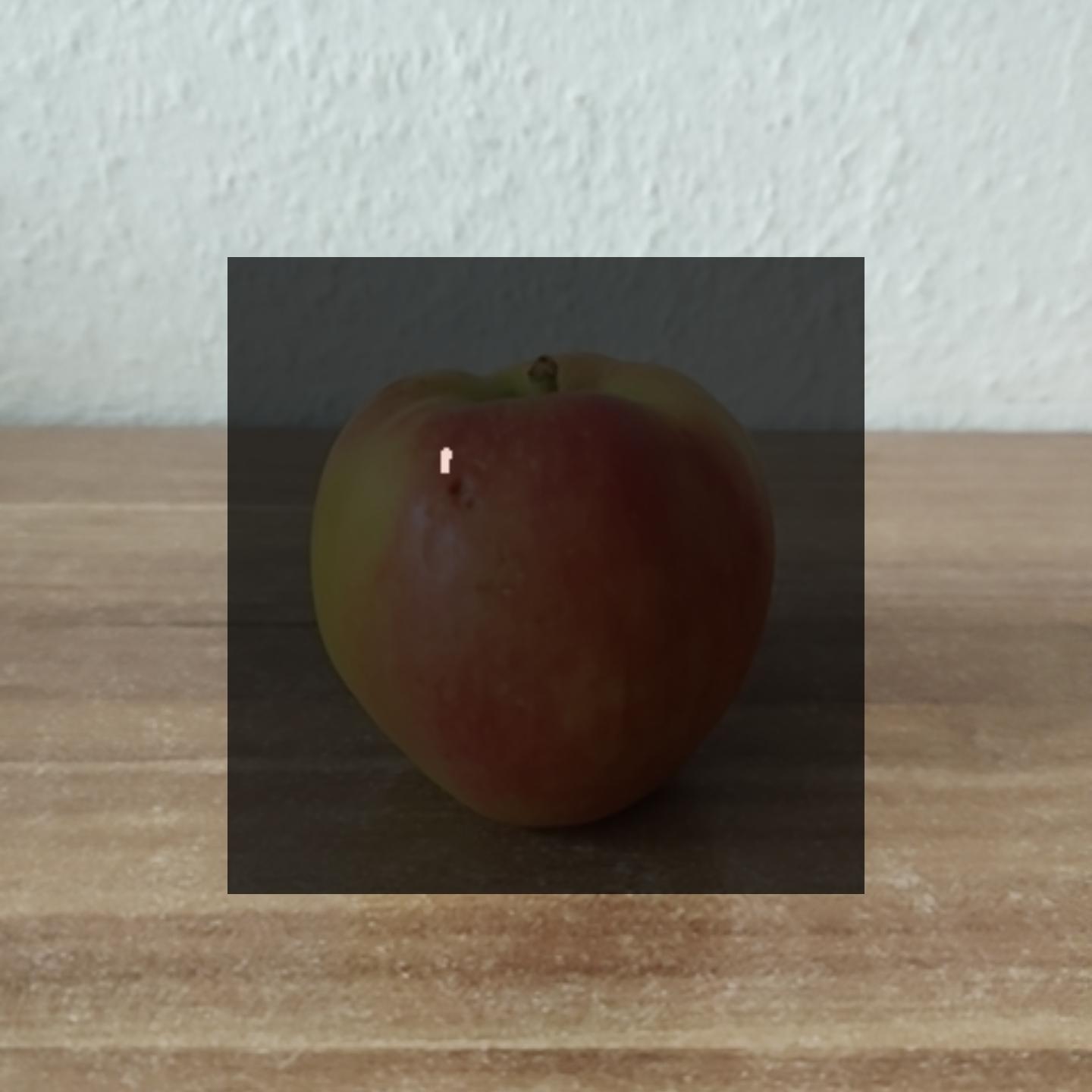}\\[4pt]
                \includegraphics[angle=90,trim=200 200 200 200, clip, width=0.9\linewidth]{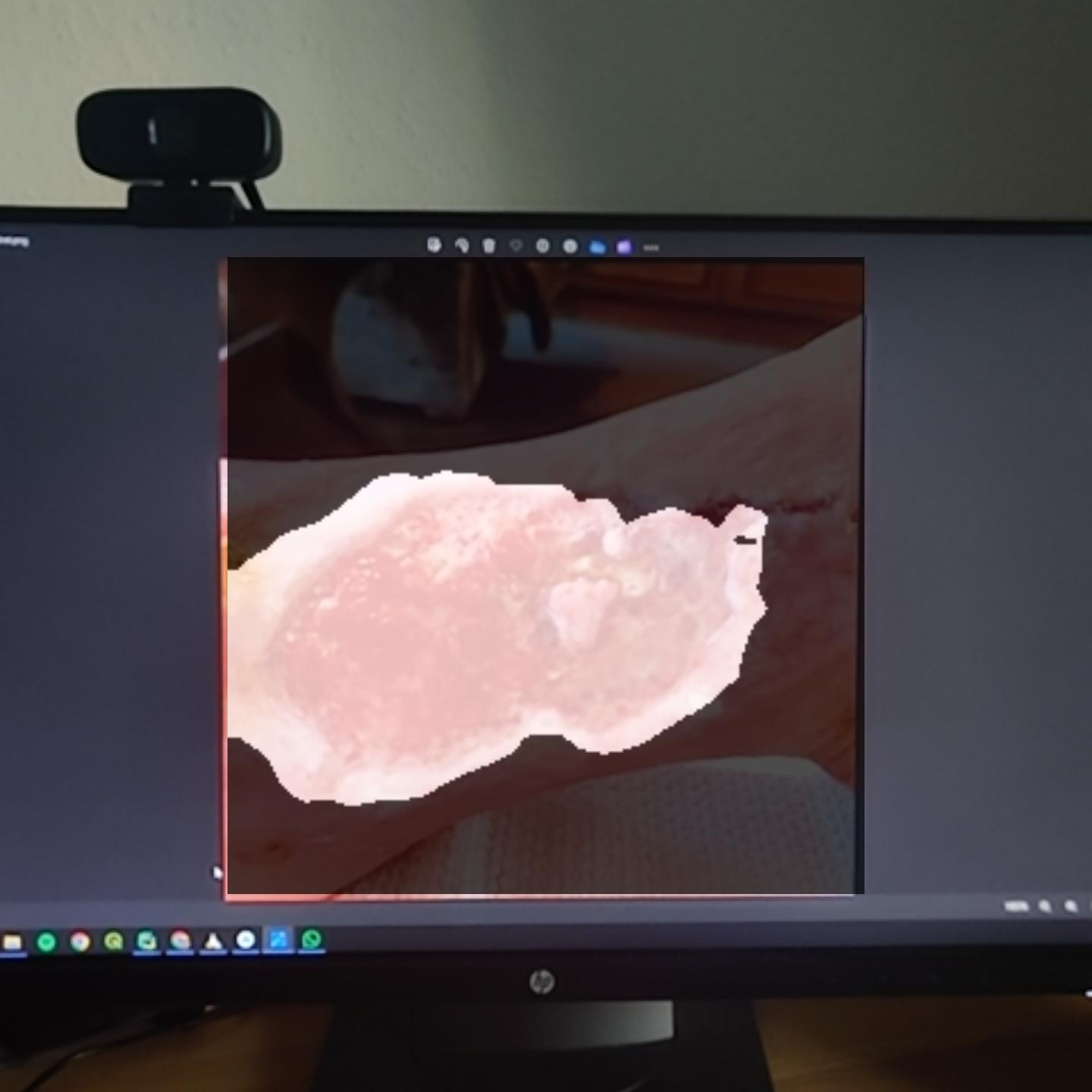}\\[4pt]
                \includegraphics[angle=90,trim=200 200 200 200, clip, width=0.9\linewidth]{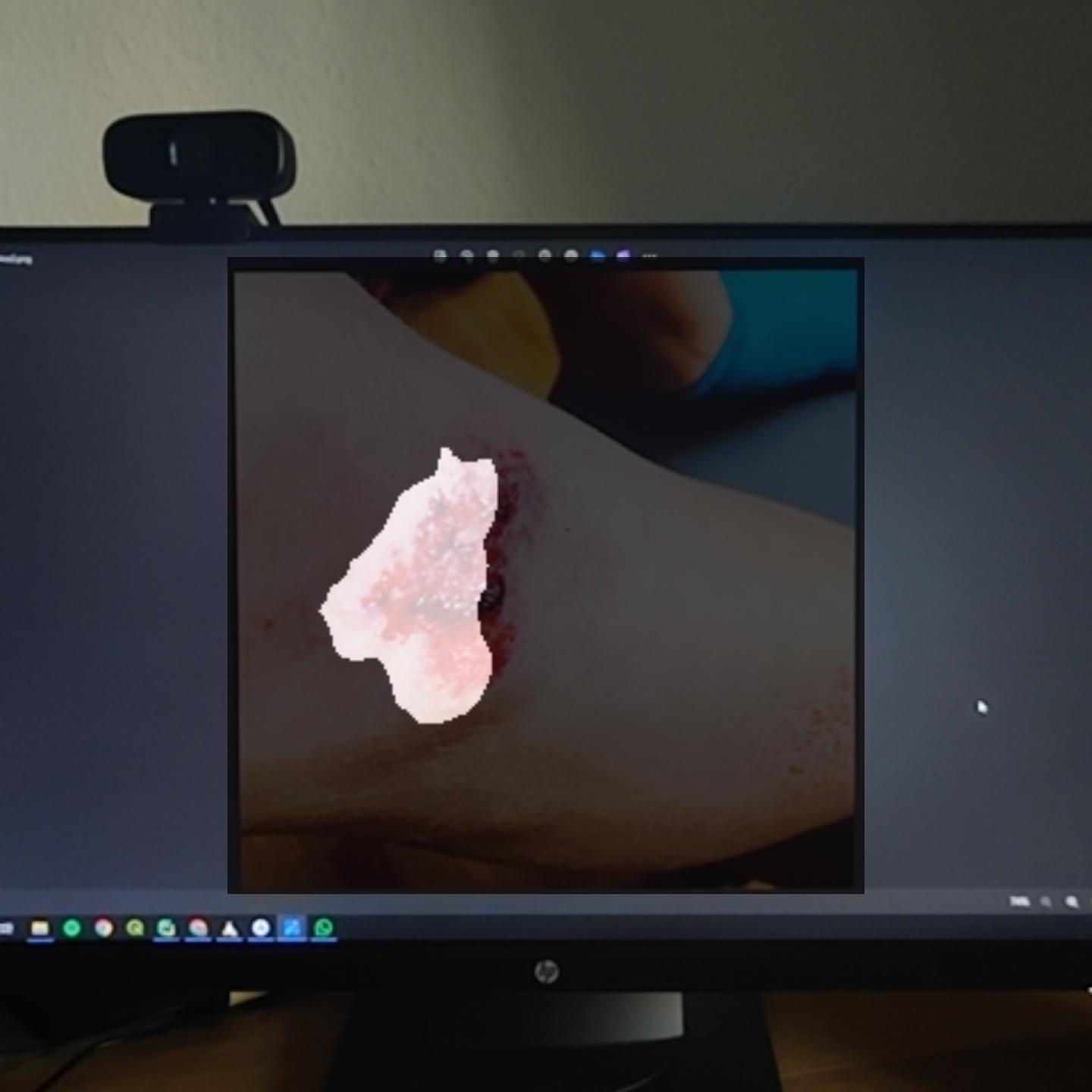}\\[4pt]
                \includegraphics[angle=90,trim=200 200 200 200, clip, width=0.9\linewidth]{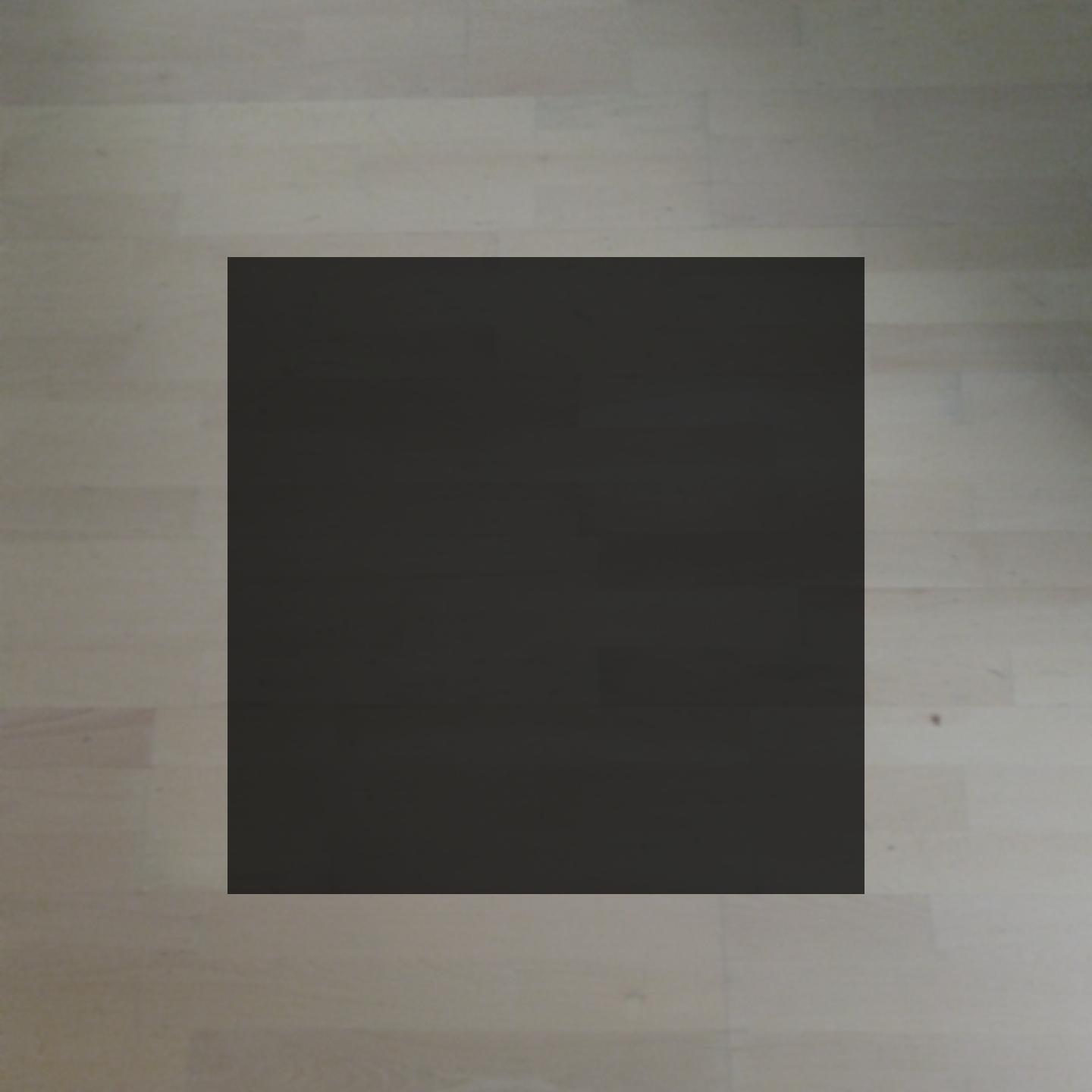}\\[4pt]
                \caption*{\scriptsize TopFormer-B}
            \end{subfigure}
            \begin{subfigure}{0.165\textwidth}
                \centering
                \includegraphics[angle=90,trim=200 200 200 200, clip, width=0.9\linewidth]{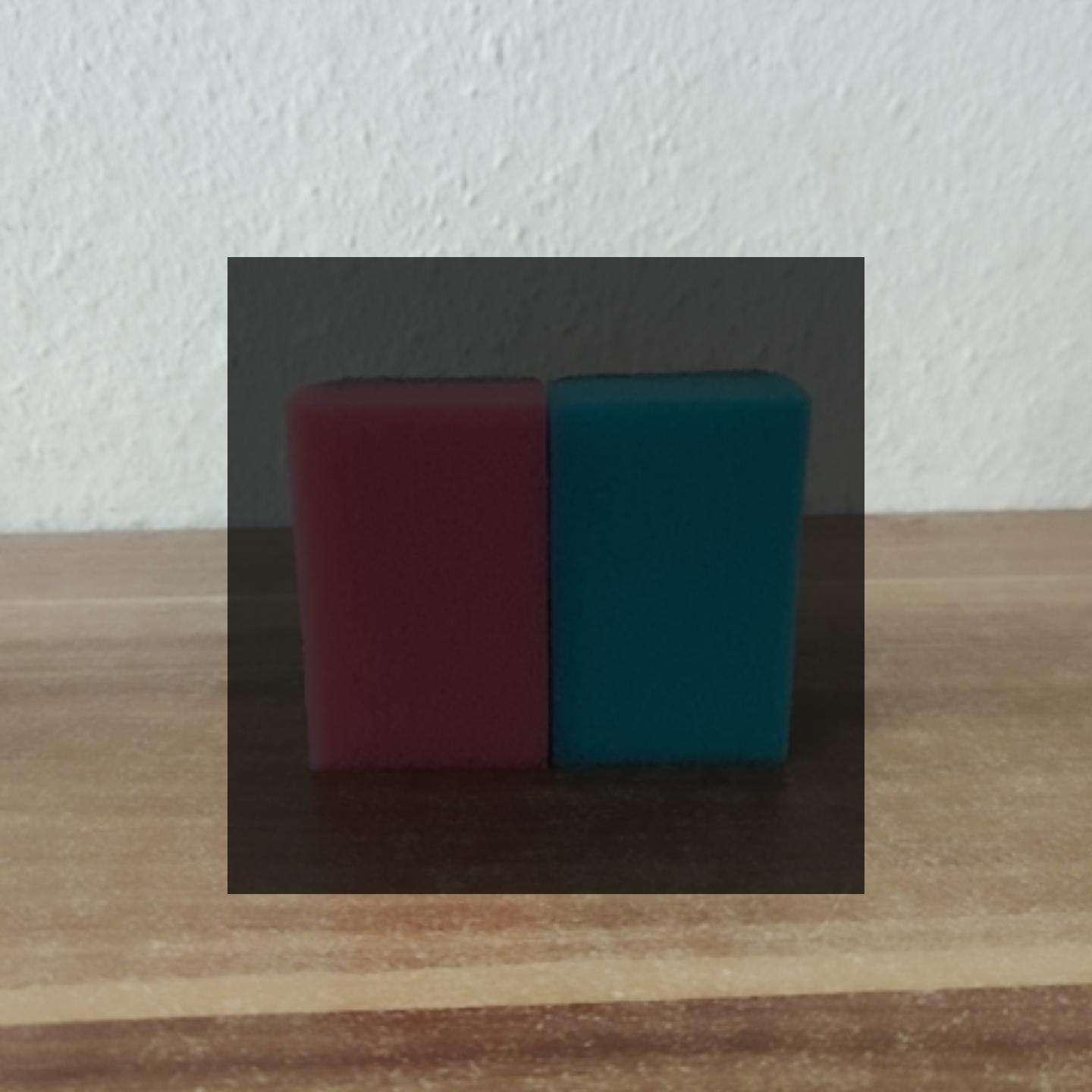}\\[4pt]
                \includegraphics[angle=90,trim=200 200 200 200, clip, width=0.9\linewidth]{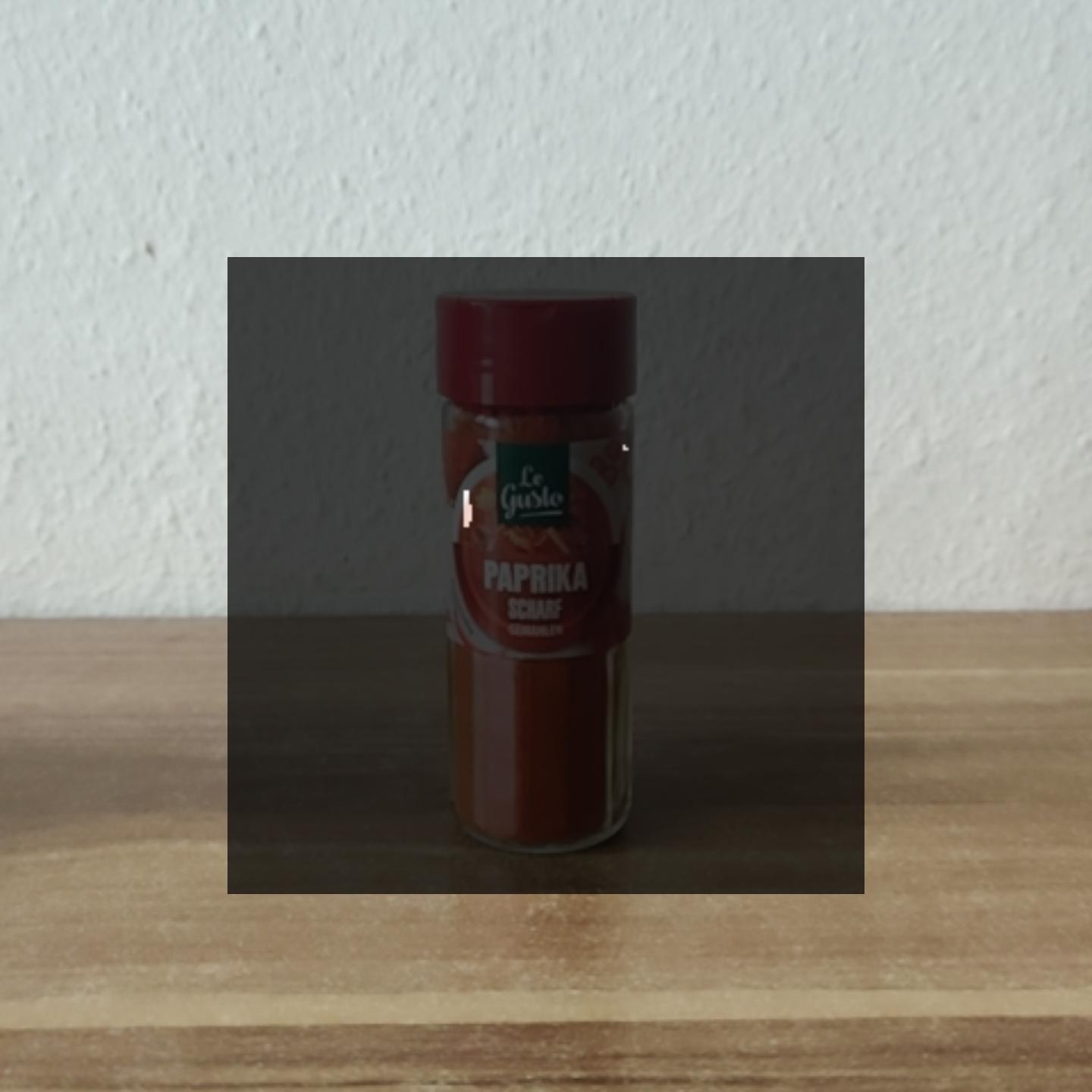}\\[4pt]
                \includegraphics[angle=90,trim=200 200 200 200, clip, width=0.9\linewidth]{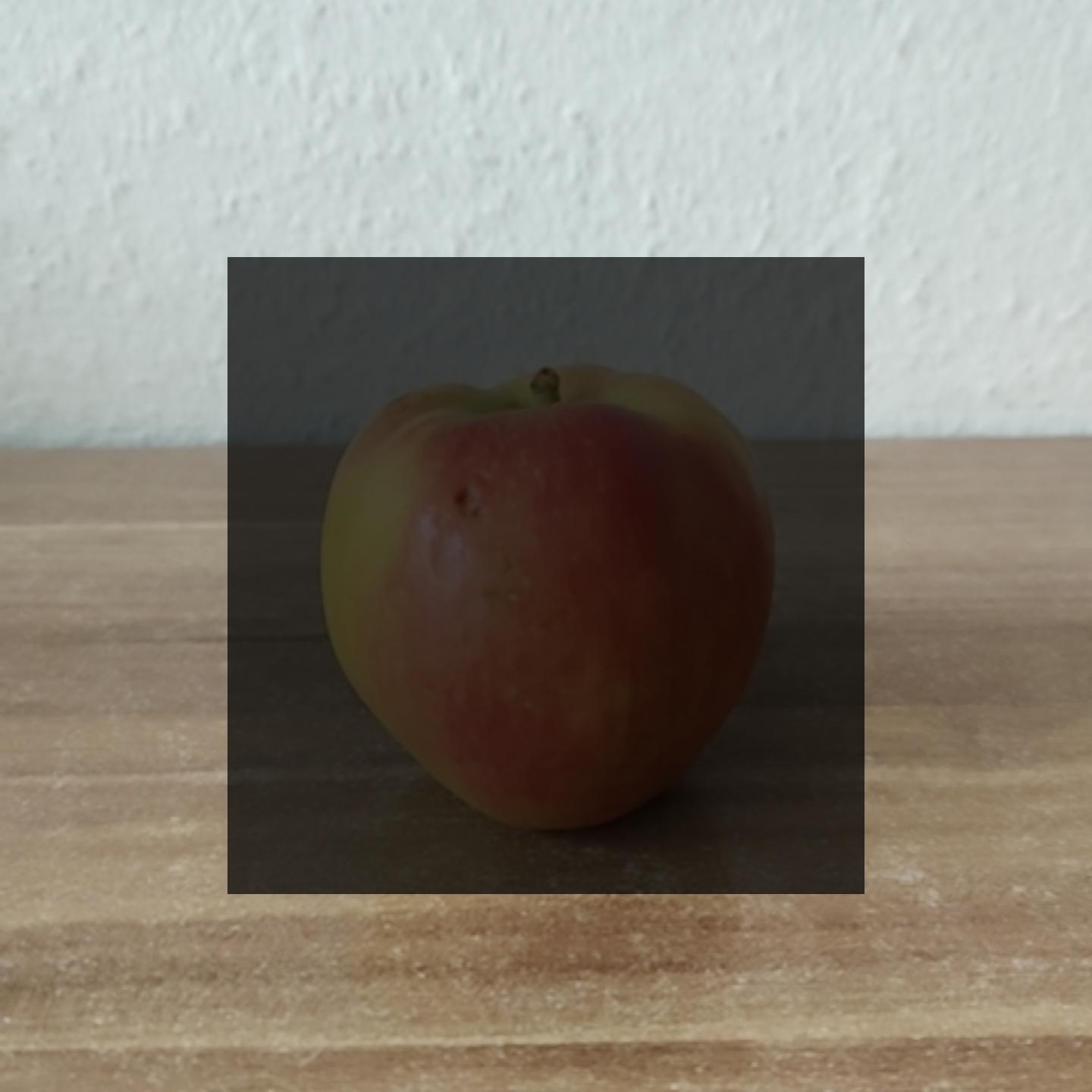}\\[4pt]
                \includegraphics[angle=90,trim=200 200 200 200, clip, width=0.9\linewidth]{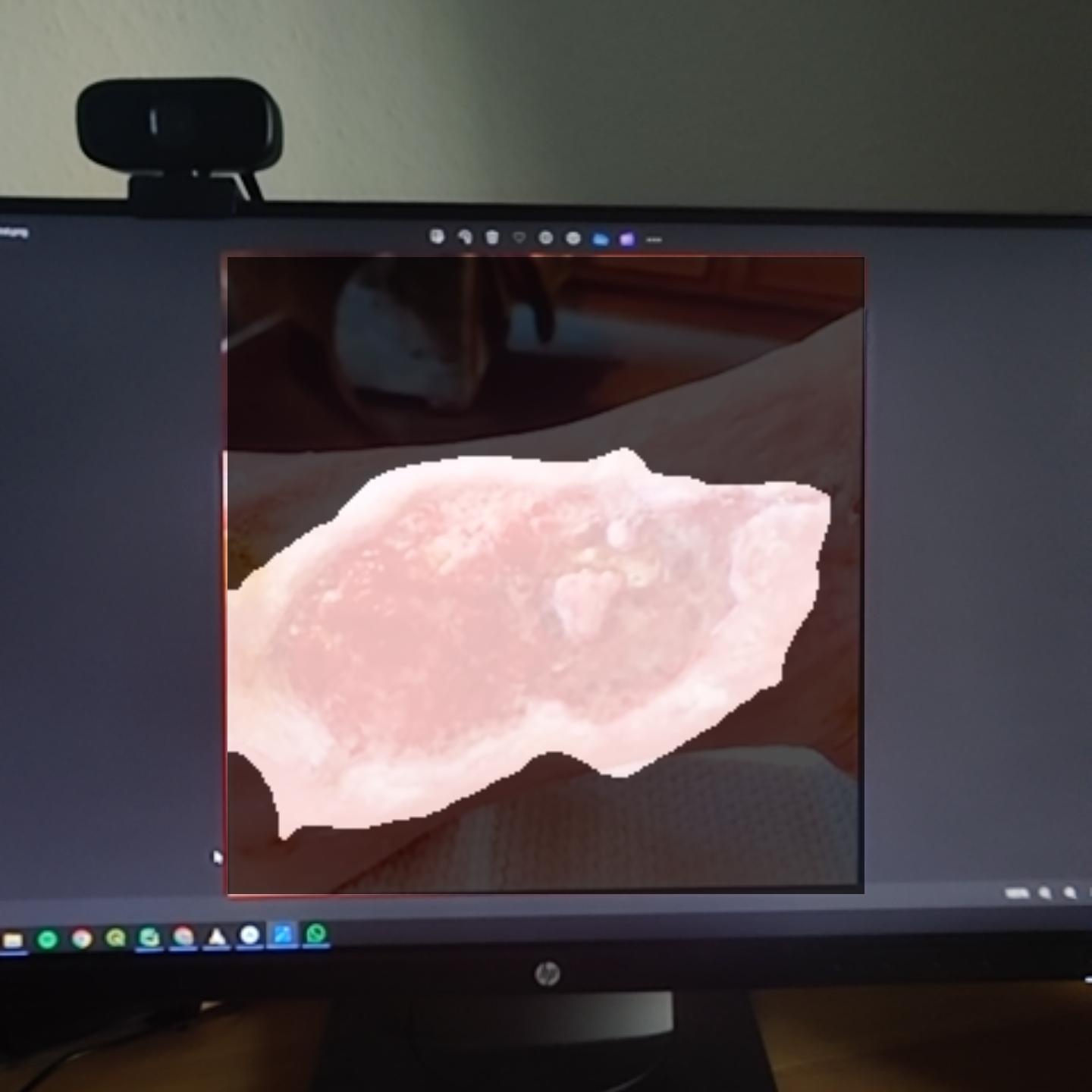}\\[4pt]
                \includegraphics[angle=90,trim=200 200 200 200, clip, width=0.9\linewidth]{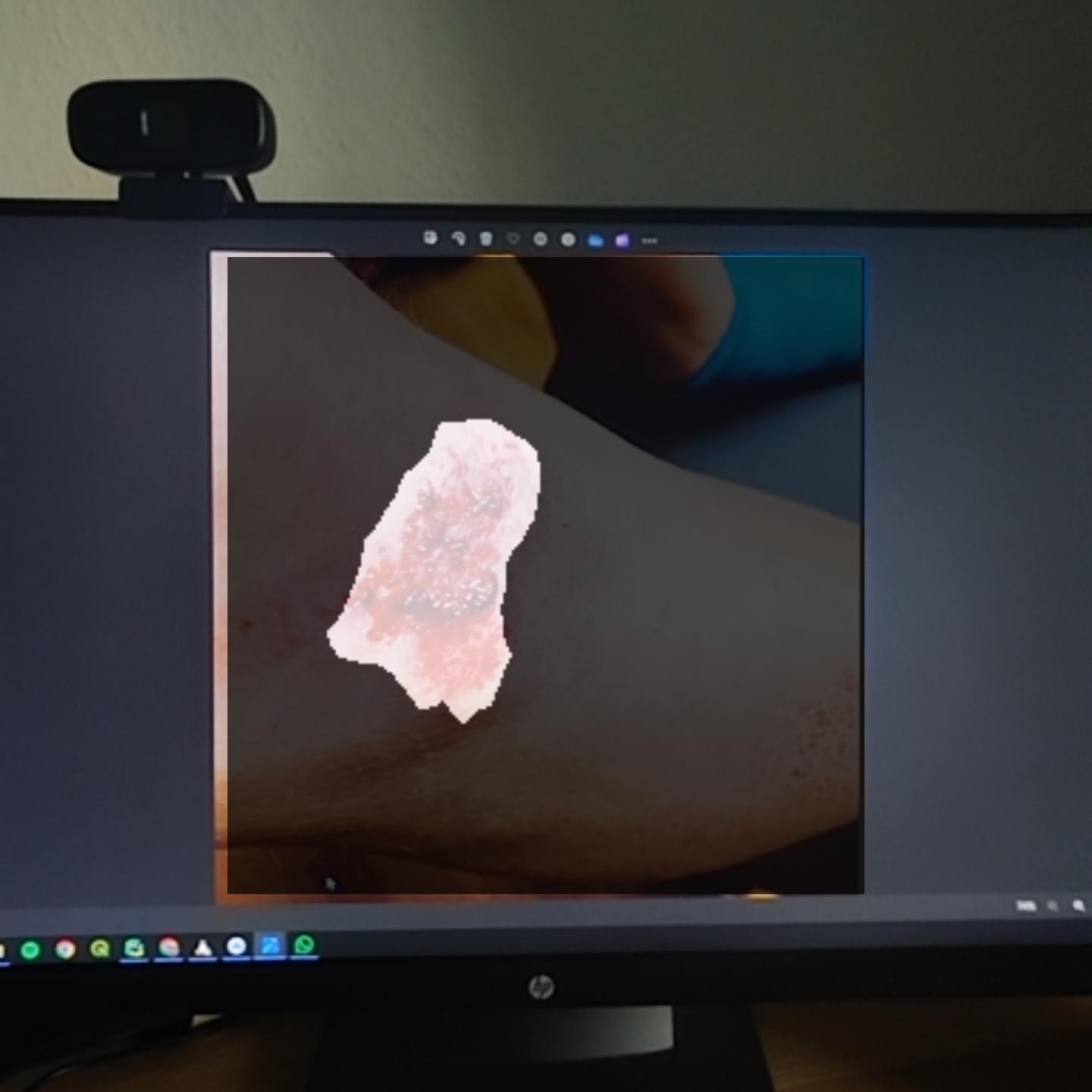}\\[4pt]
                \includegraphics[angle=90,trim=200 200 200 200, clip, width=0.9\linewidth]{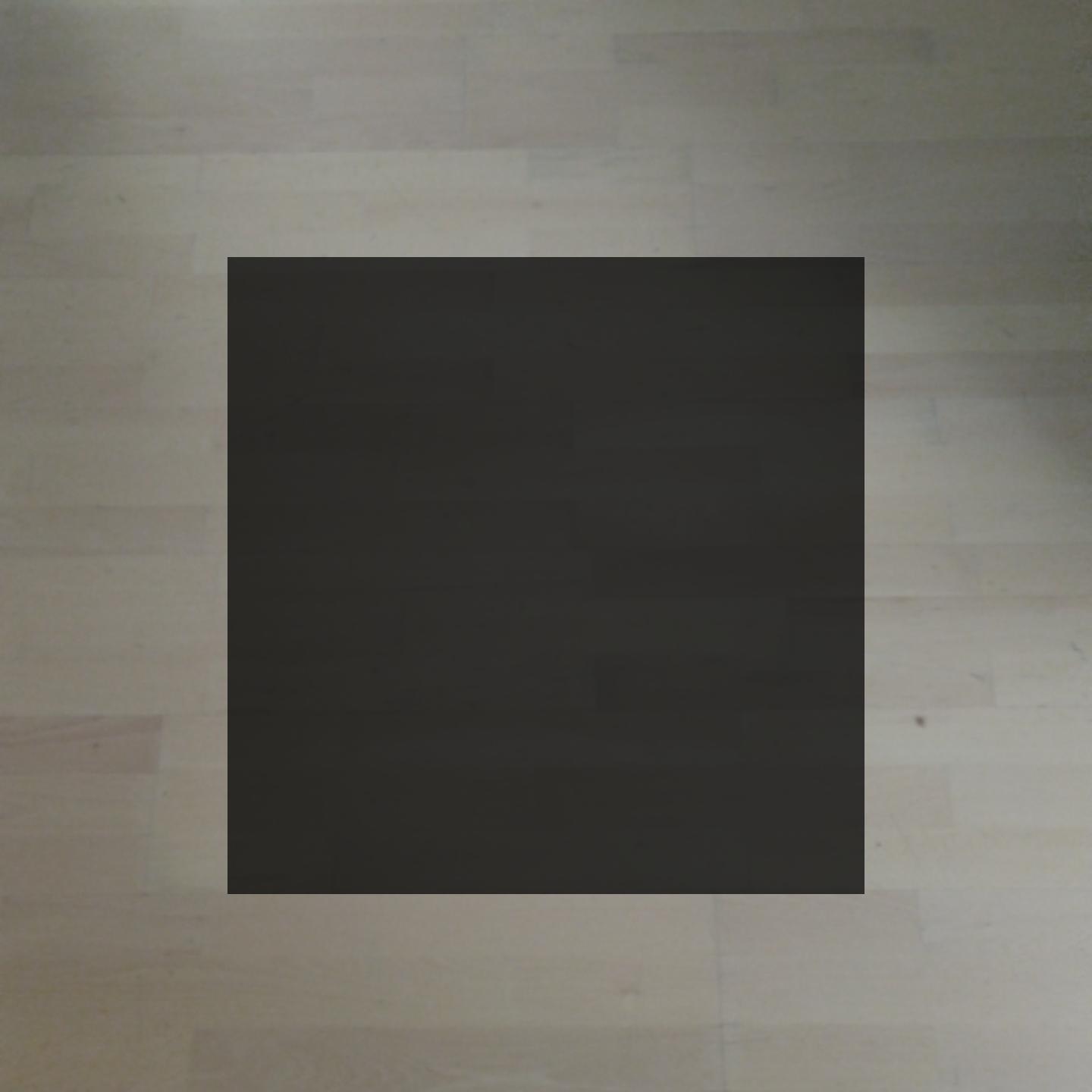}\\[4pt]
                \caption*{\scriptsize TopFormer-S}
            \end{subfigure}
            \begin{subfigure}{0.165\textwidth}
                \centering
                \includegraphics[angle=90,trim=200 200 200 200, clip, width=0.9\linewidth]{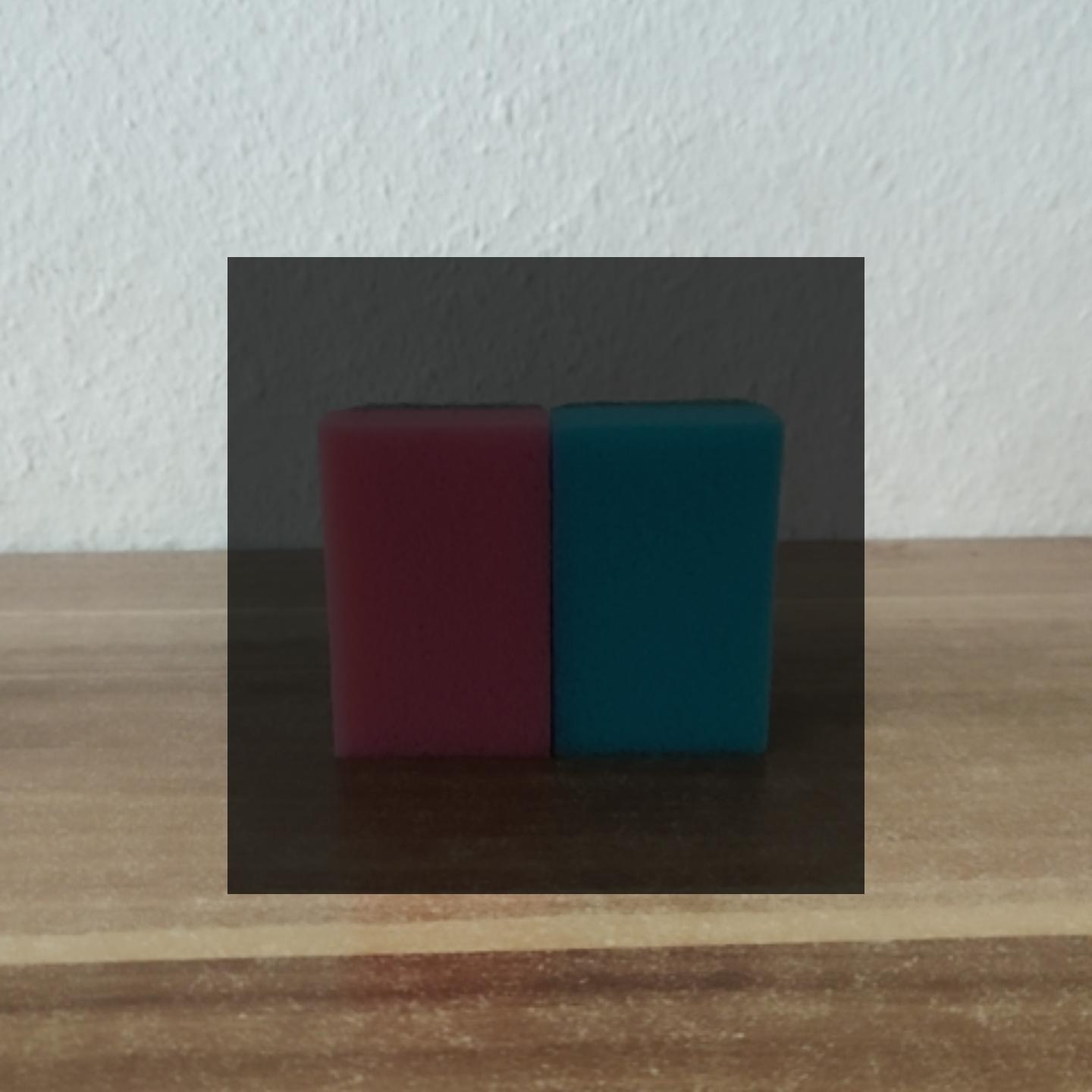}\\[4pt]
                \includegraphics[angle=90,trim=200 200 200 200, clip, width=0.9\linewidth]{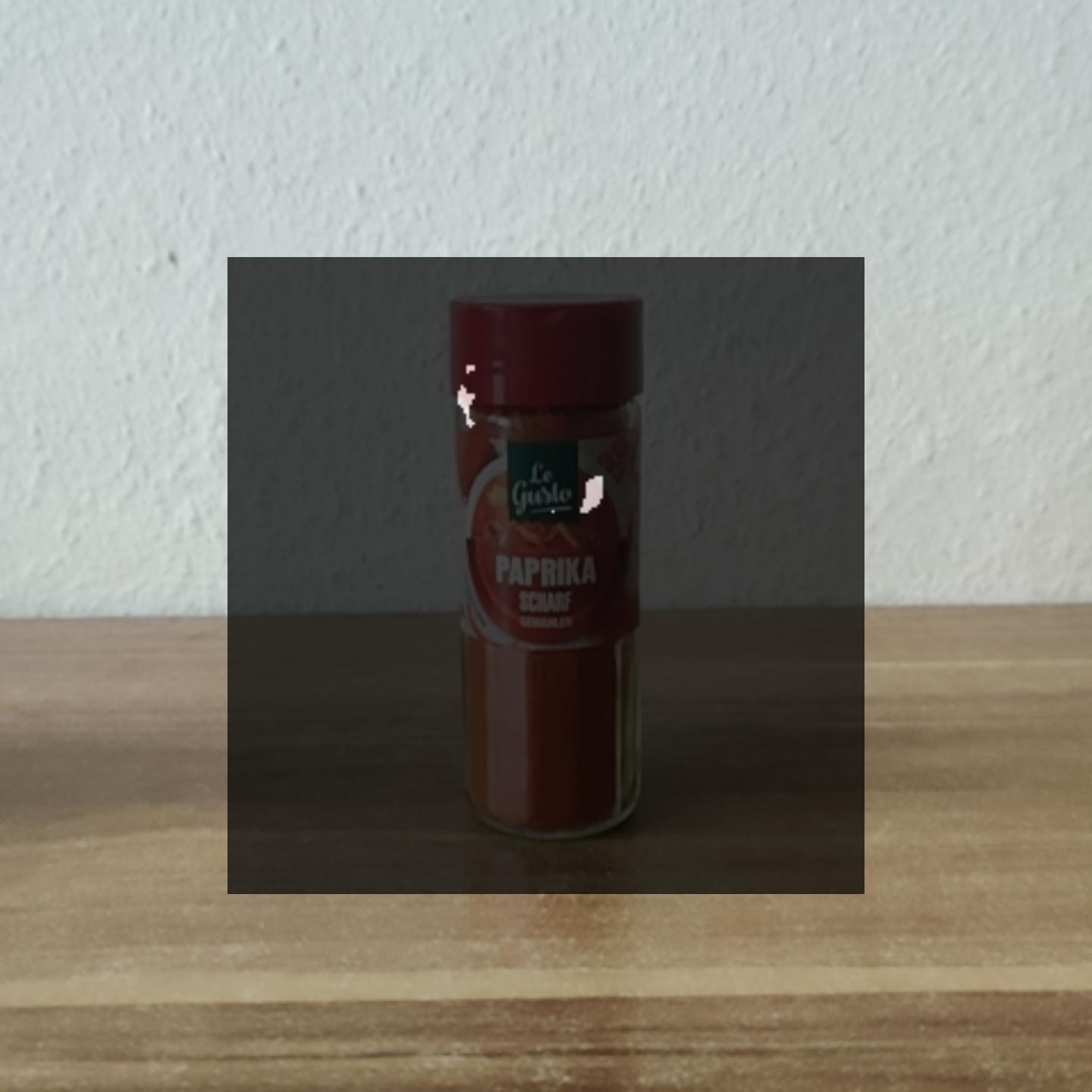}\\[4pt]
                \includegraphics[angle=90,trim=200 200 200 200, clip, width=0.9\linewidth]{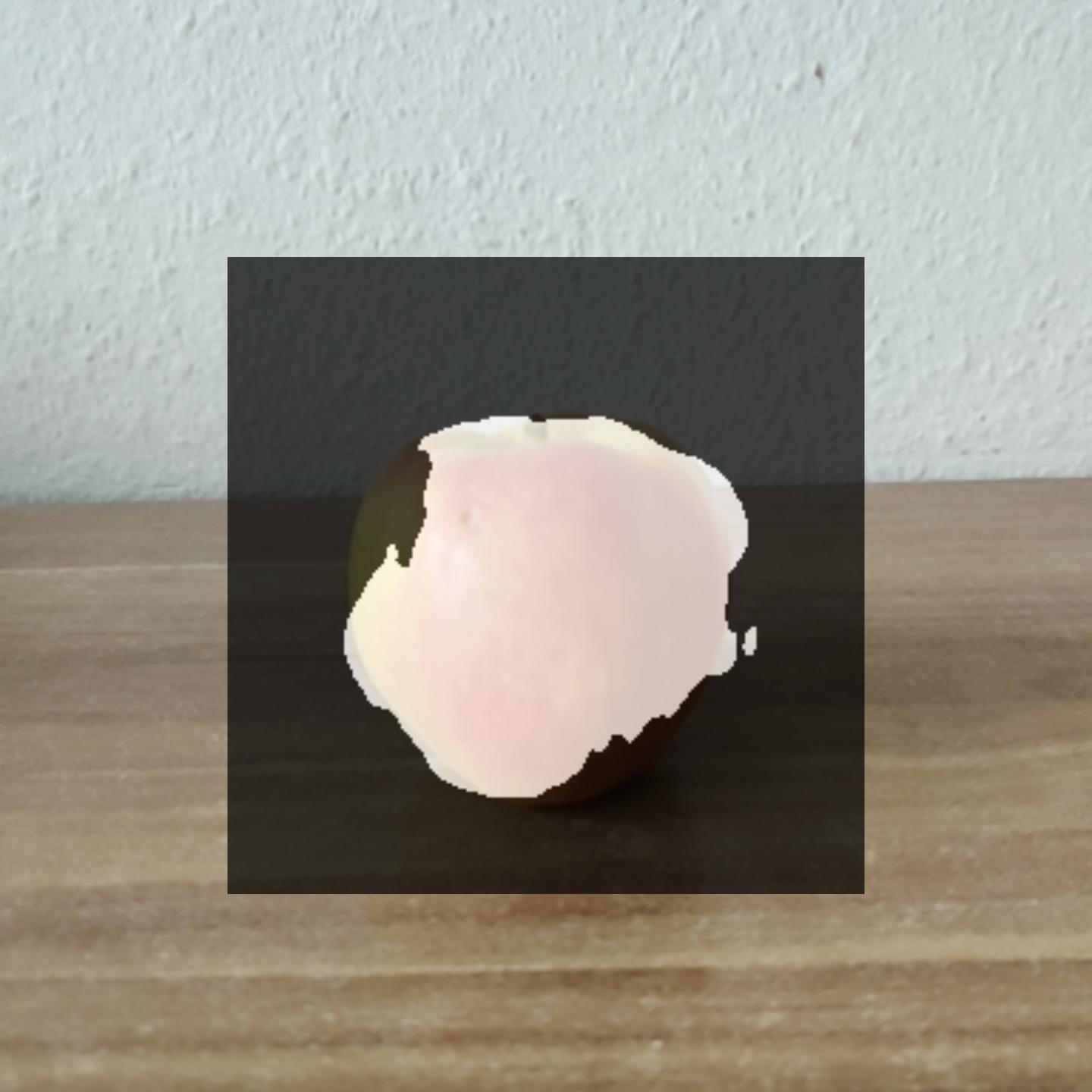}\\[4pt]
                \includegraphics[angle=90,trim=200 200 200 200, clip, width=0.9\linewidth]{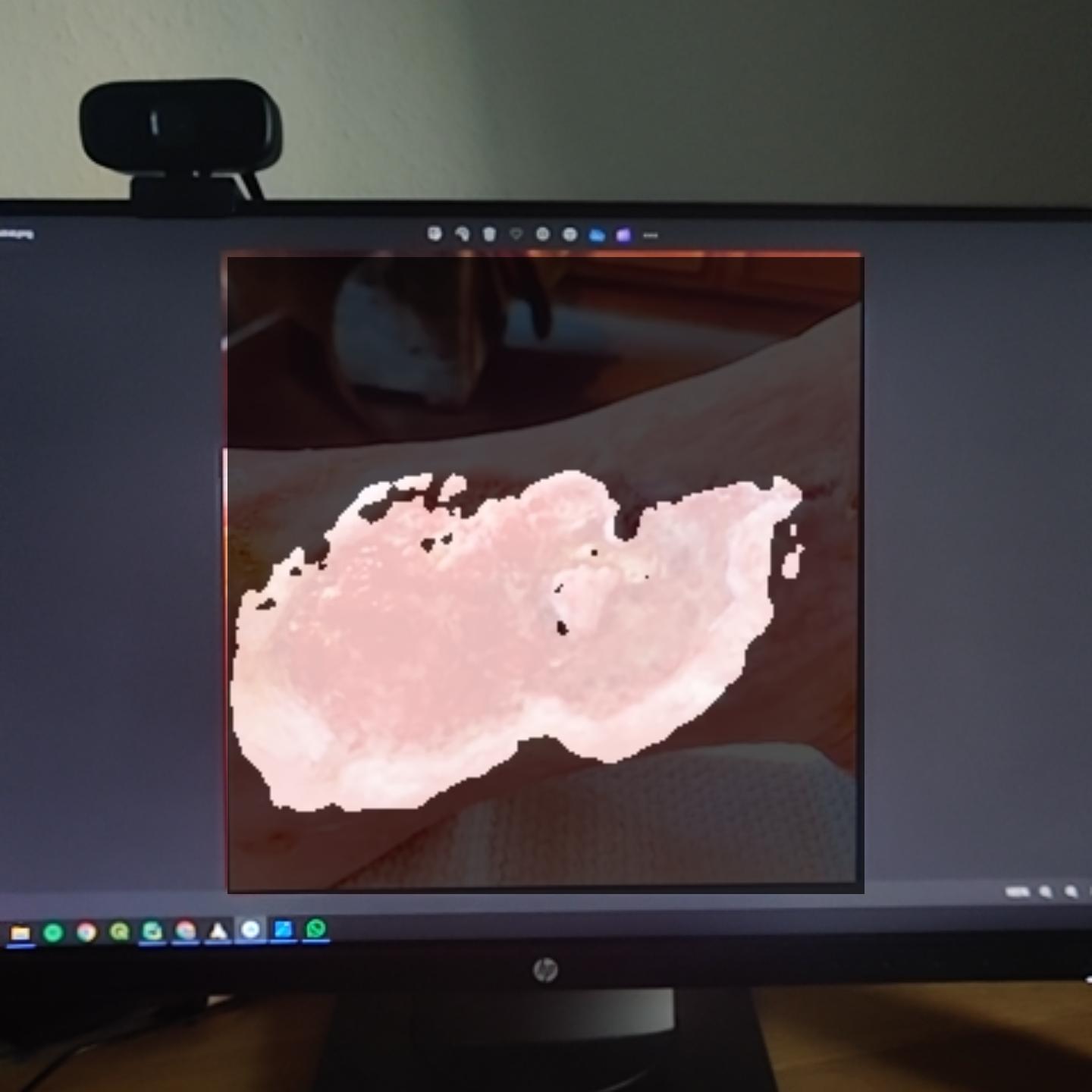}\\[4pt]
                \includegraphics[angle=90,trim=200 200 200 200, clip, width=0.9\linewidth]{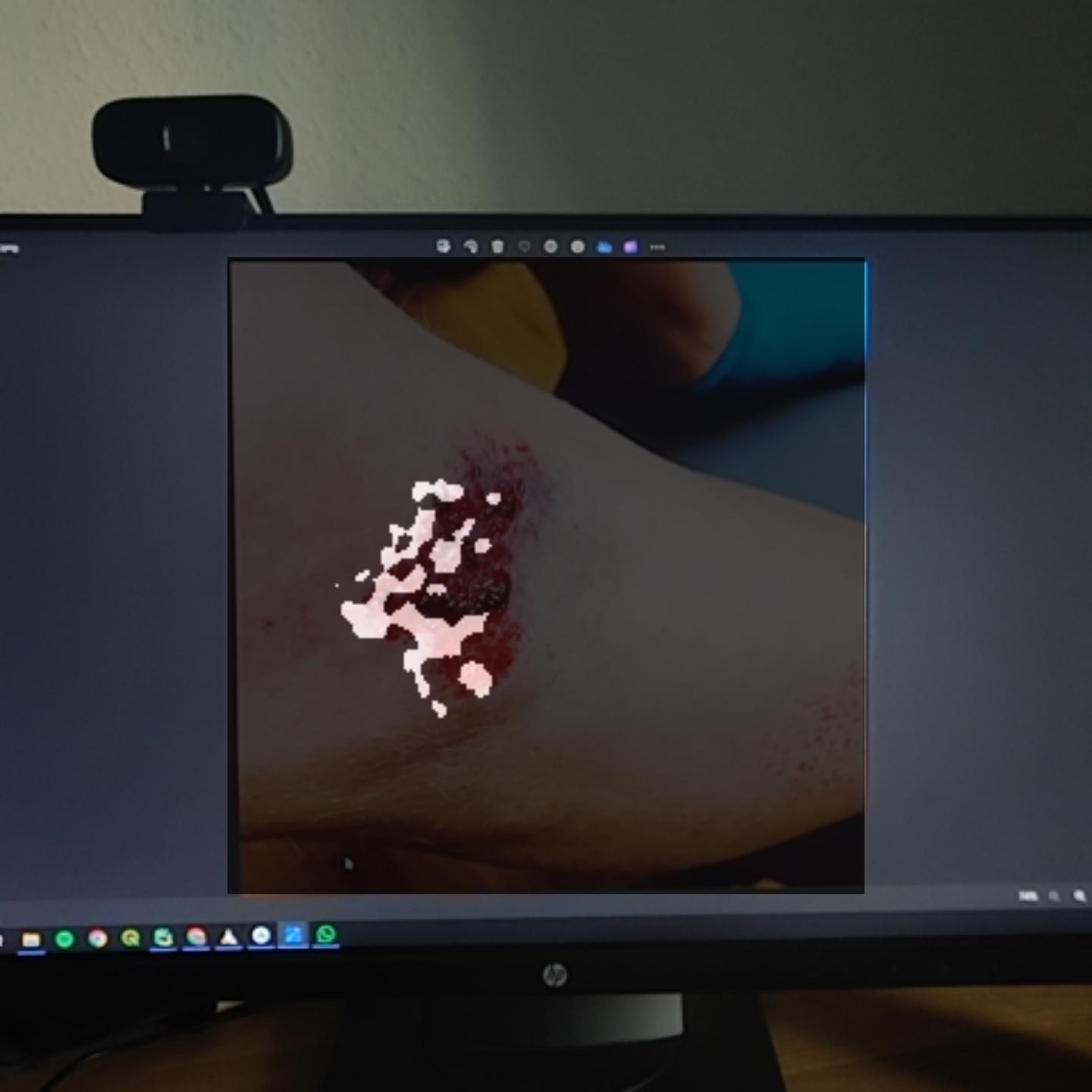}\\[4pt]
                \includegraphics[angle=90,trim=200 200 200 200, clip, width=0.9\linewidth]{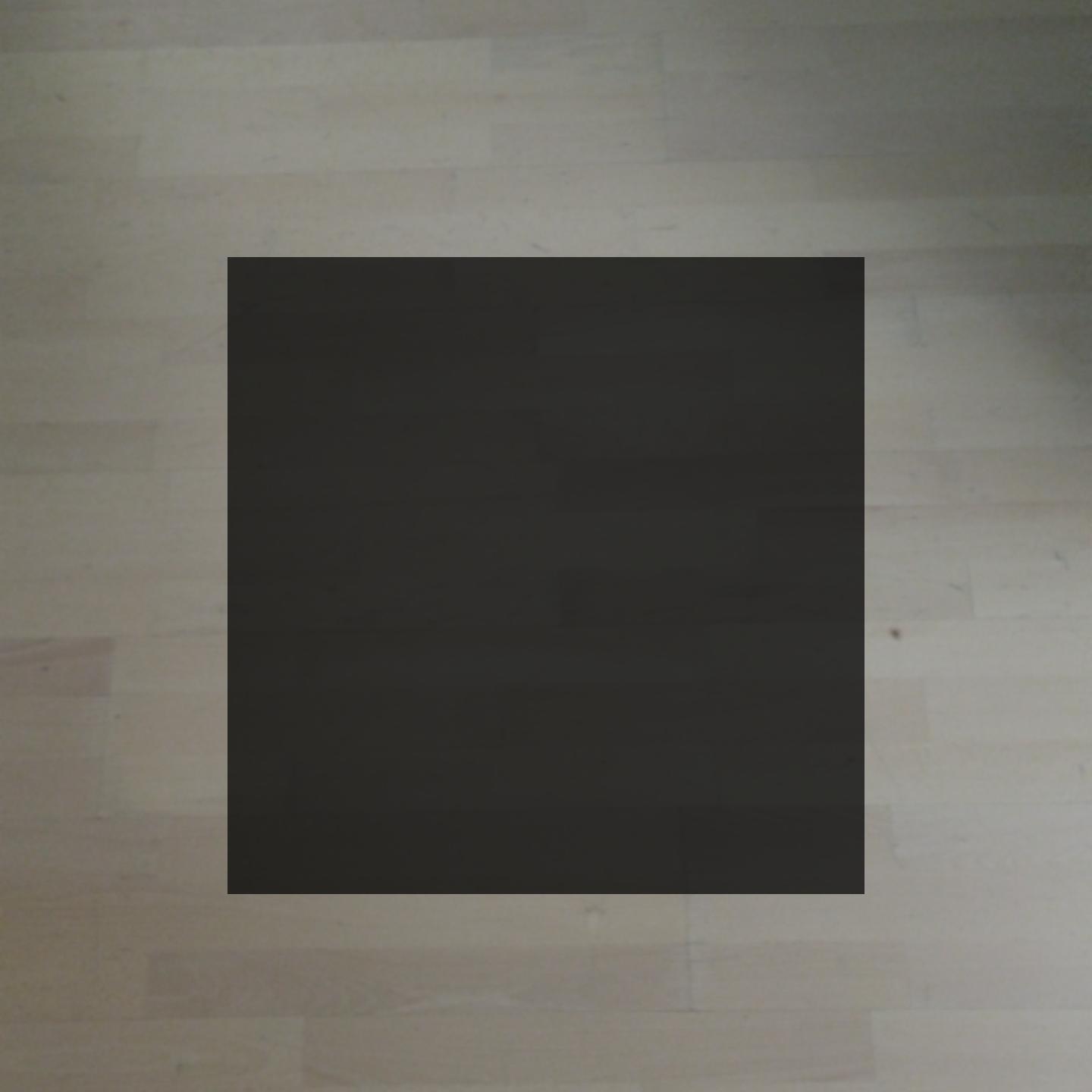}\\[4pt]
                \caption*{\scriptsize UNeXt-B}
            \end{subfigure}
            \begin{subfigure}{0.165\textwidth}
                \centering
                \includegraphics[angle=90,trim=200 200 200 200, clip, width=0.9\linewidth]{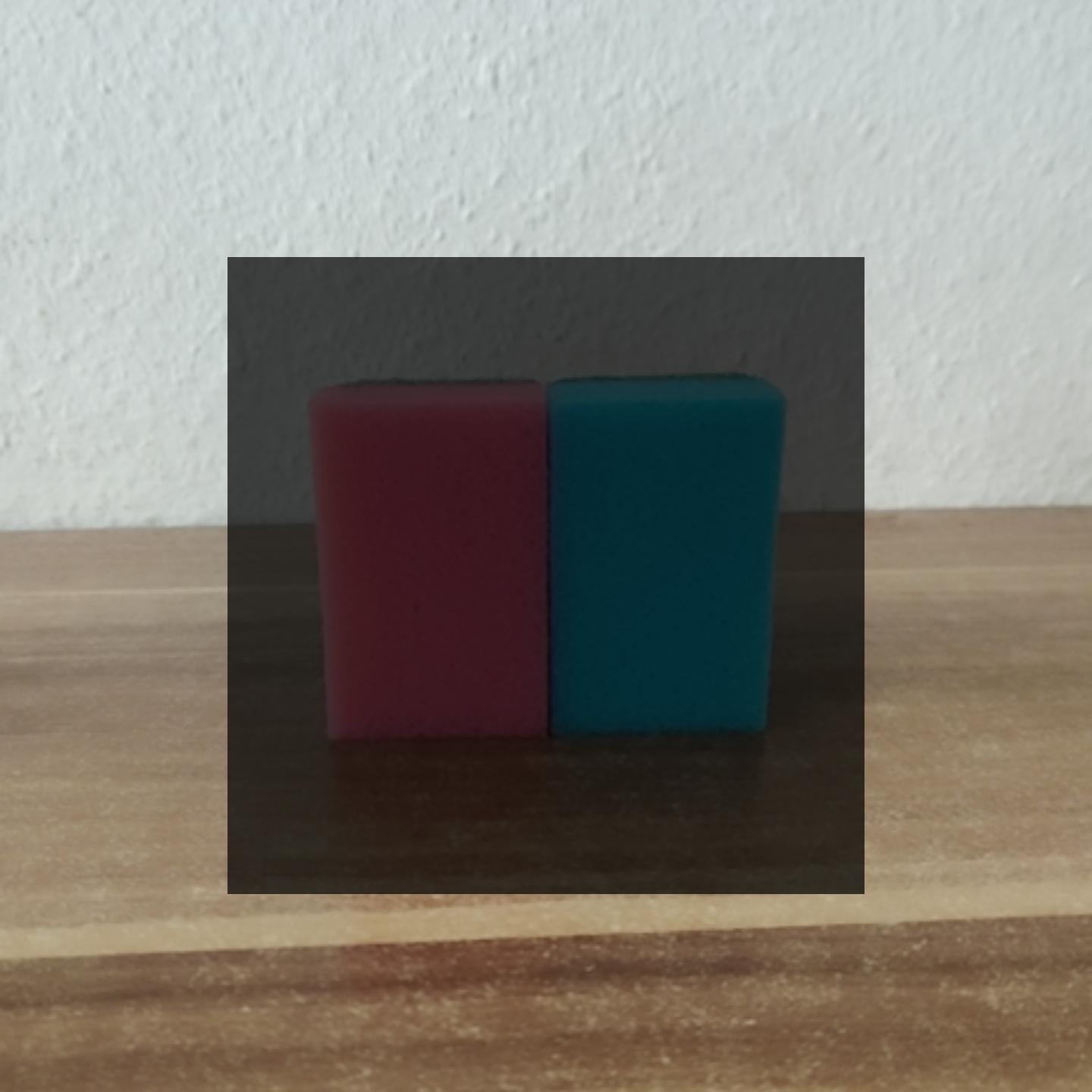}\\[4pt]
                \includegraphics[angle=90,trim=200 200 200 200, clip, width=0.9\linewidth]{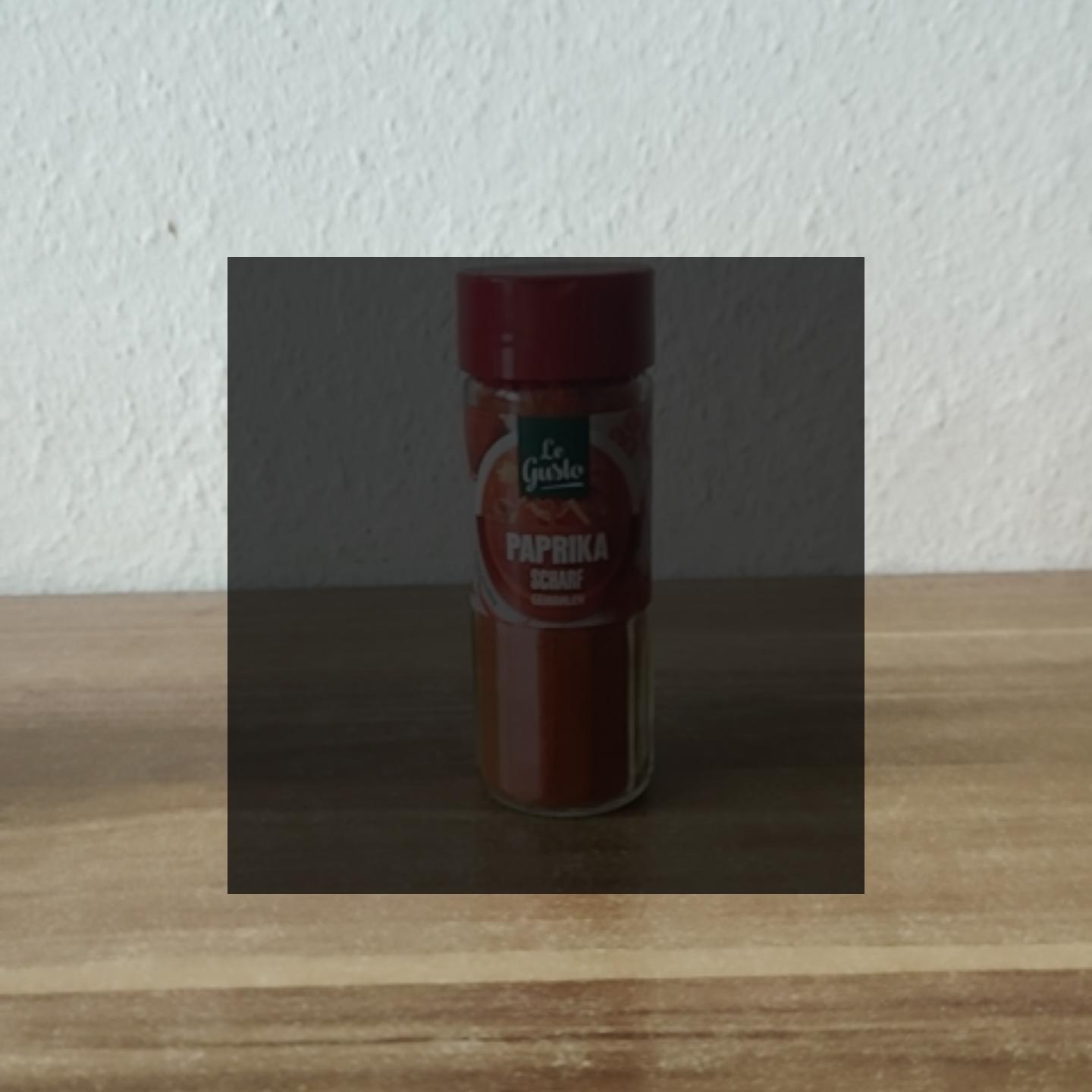}\\[4pt]
                \includegraphics[angle=90,trim=200 200 200 200, clip, width=0.9\linewidth]{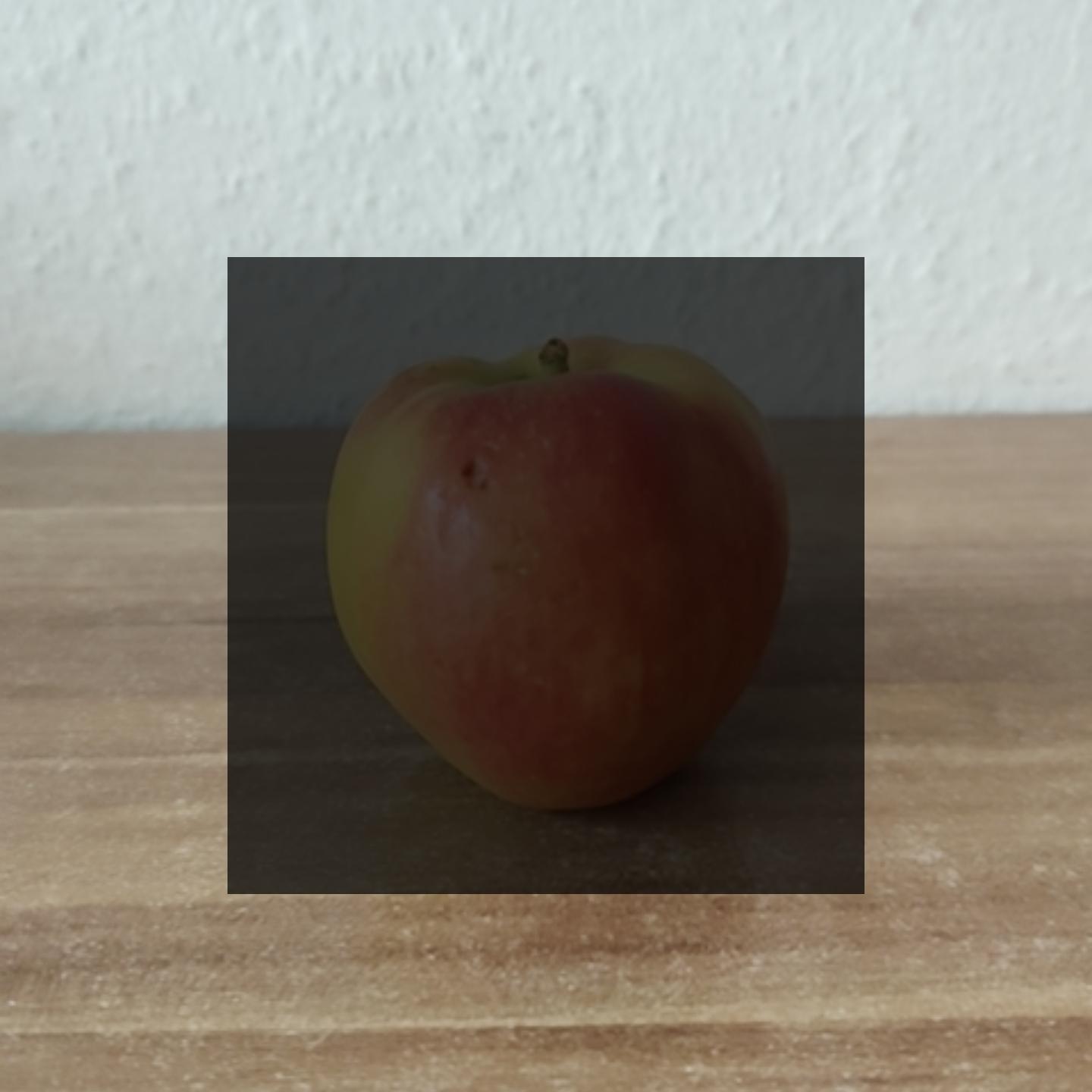}\\[4pt]
                \includegraphics[angle=90,trim=200 200 200 200, clip, width=0.9\linewidth]{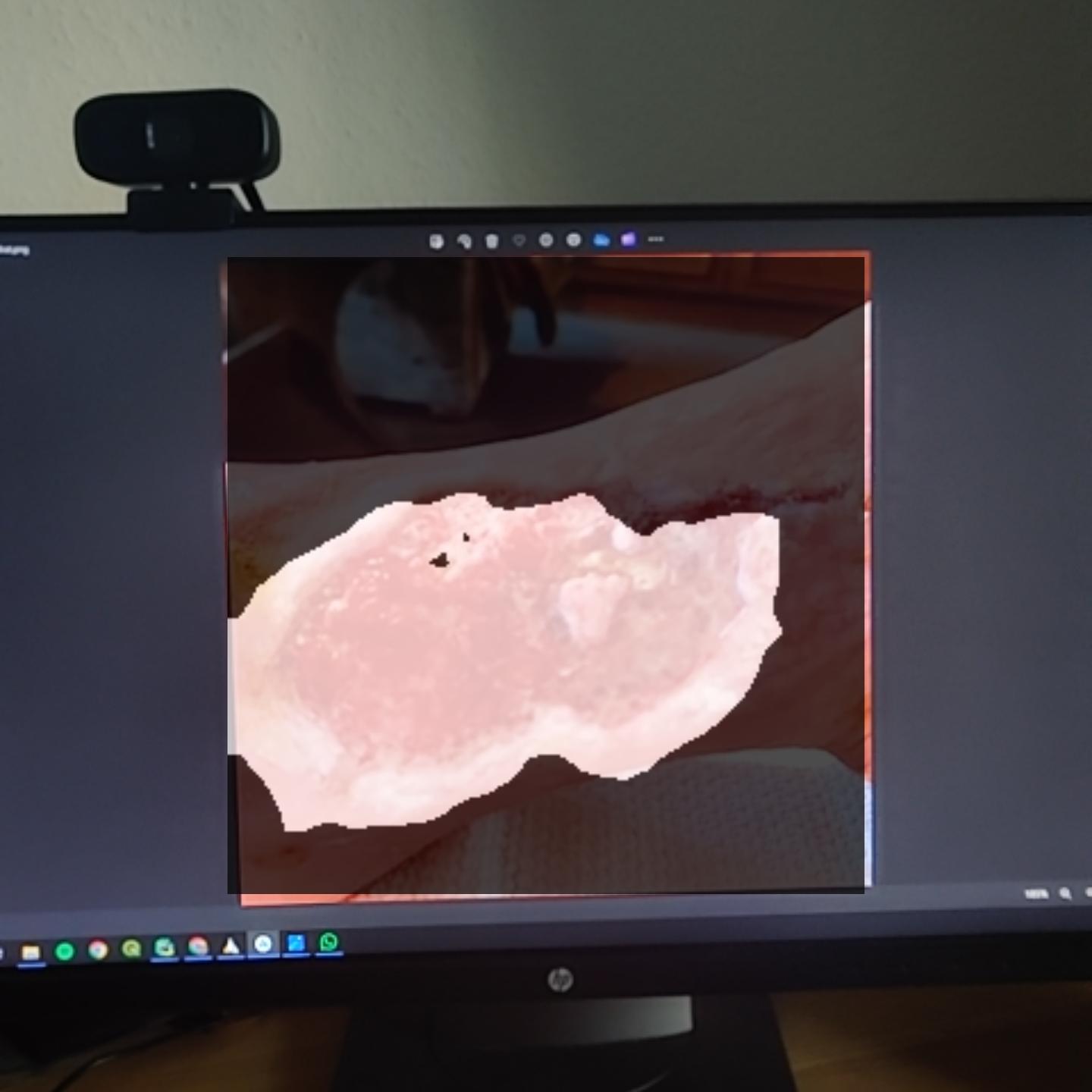}\\[4pt]
                \includegraphics[angle=90,trim=200 200 200 200, clip, width=0.9\linewidth]{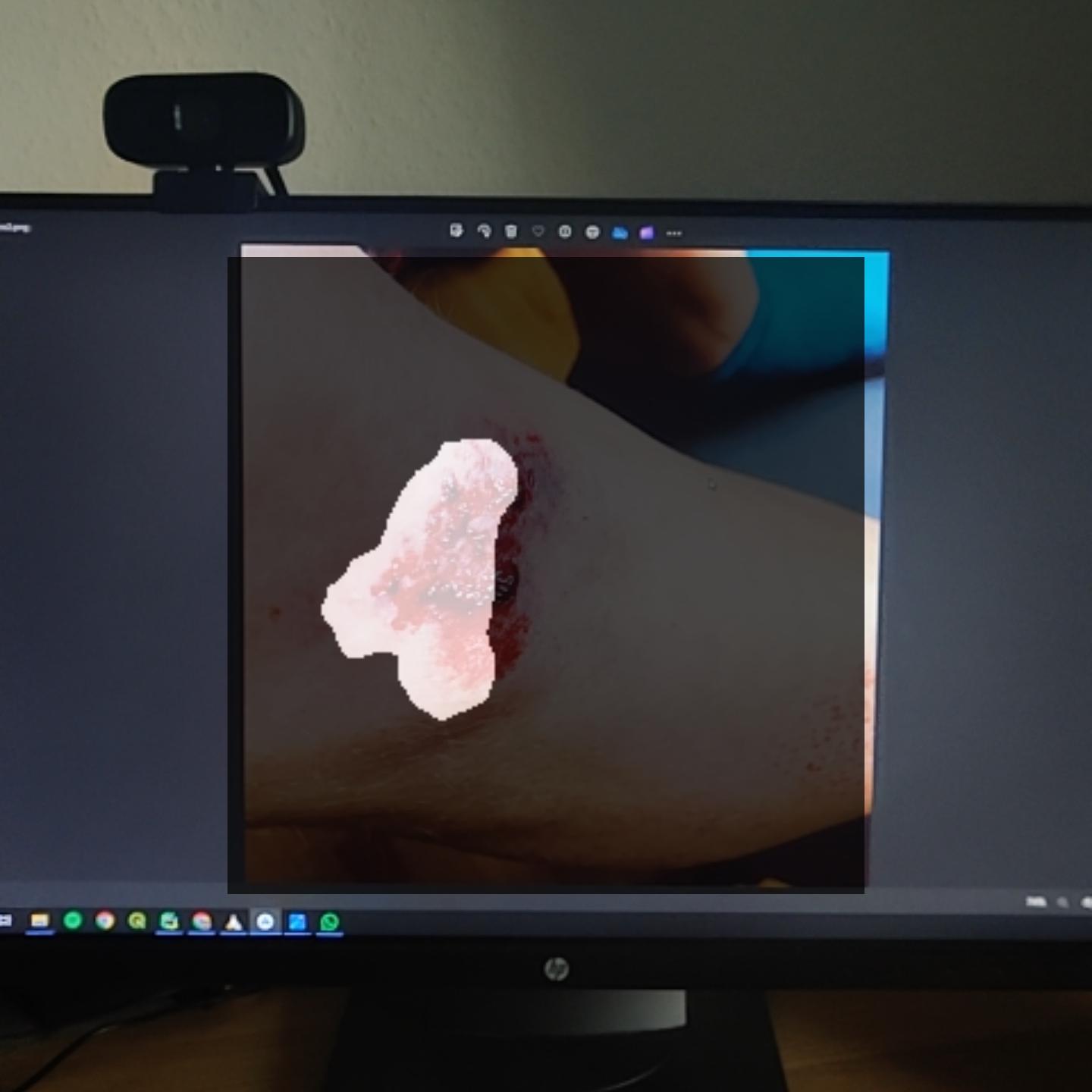}\\[4pt]
                \includegraphics[angle=90,trim=200 200 200 200, clip, width=0.9\linewidth]{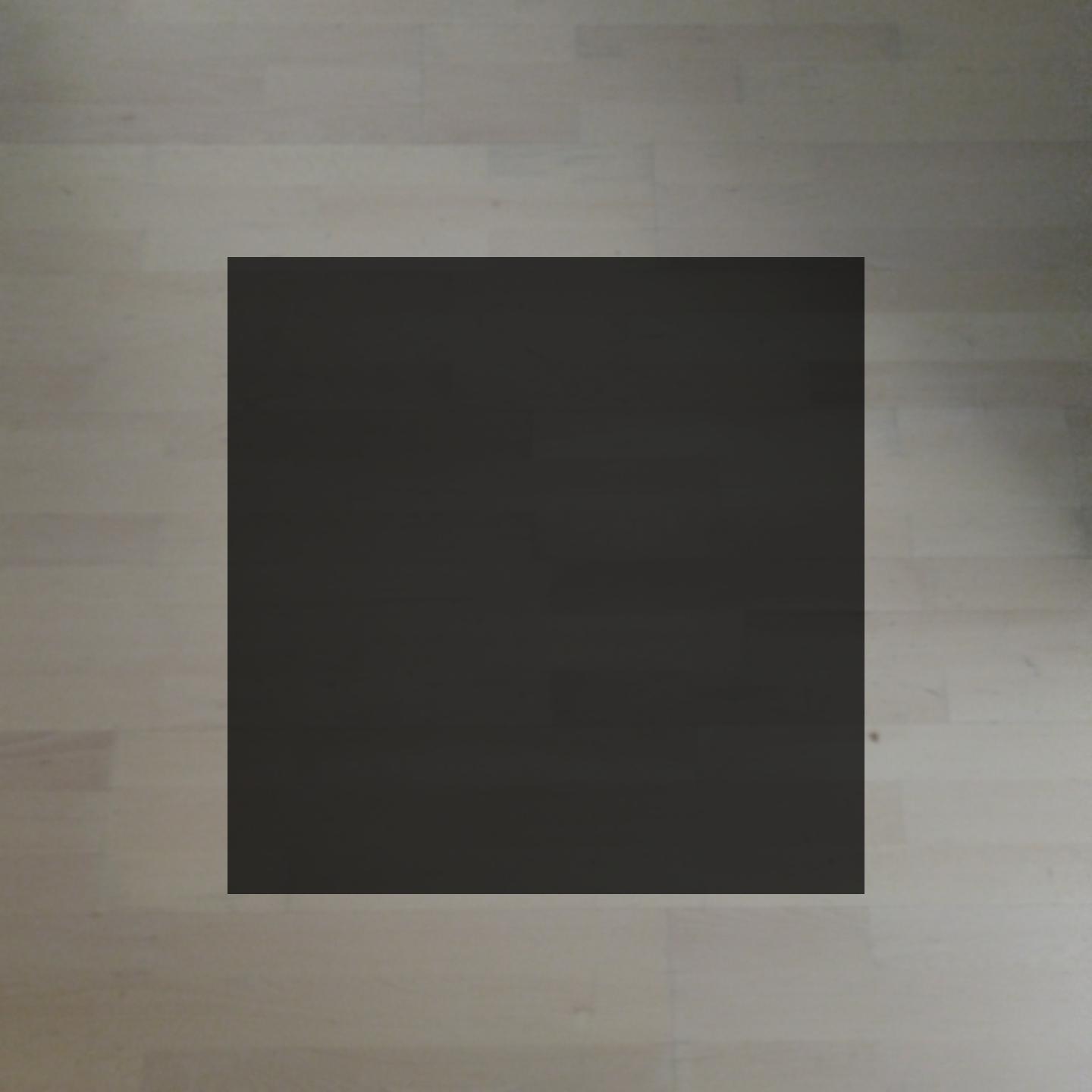}\\[4pt]
                \caption*{\scriptsize TopFormer-T}
            \end{subfigure}
            \begin{subfigure}{0.165\textwidth}
                \centering
                \includegraphics[angle=90,trim=200 200 200 200, clip, width=0.9\linewidth]{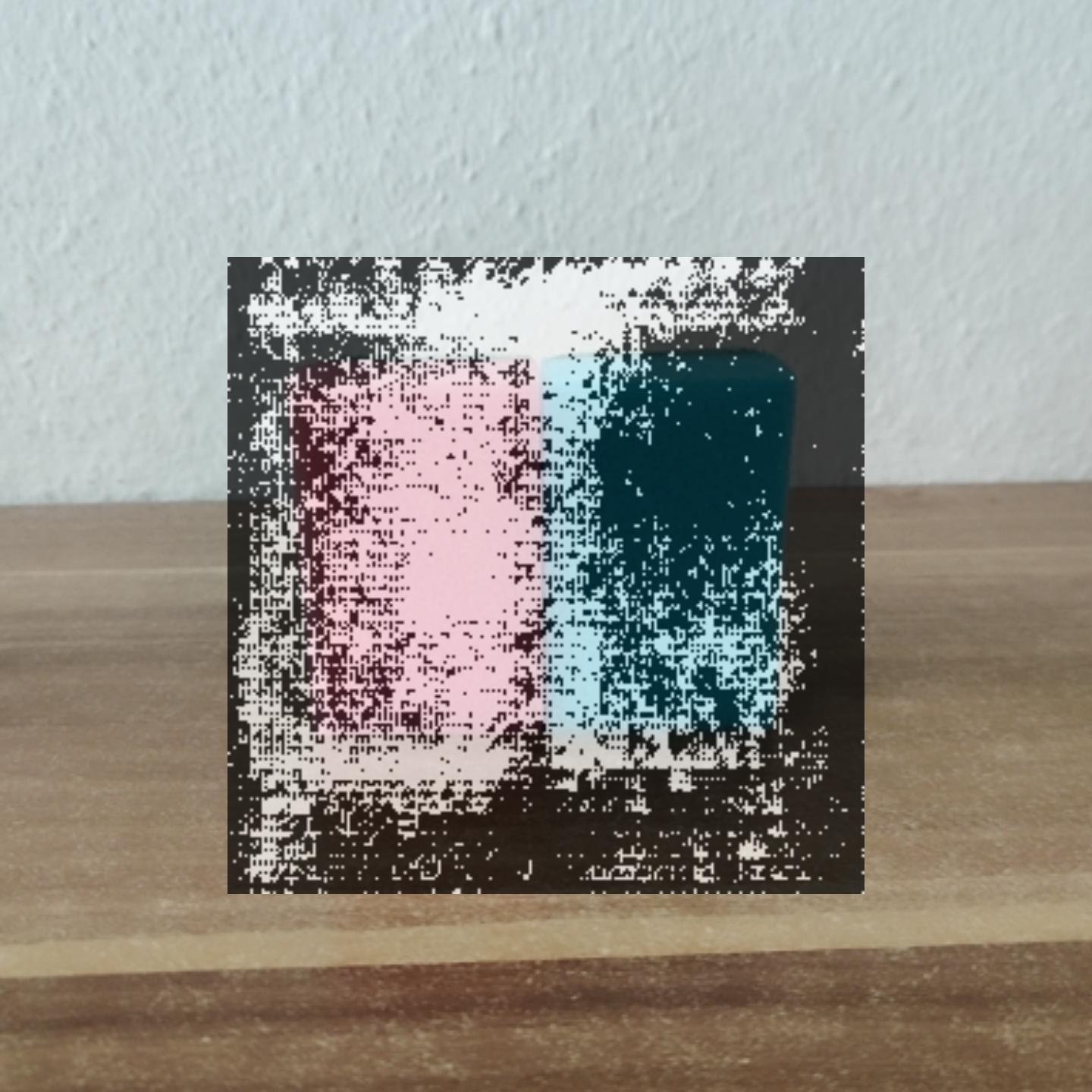}\\[4pt]
                \includegraphics[angle=90,trim=200 200 200 200, clip, width=0.9\linewidth]{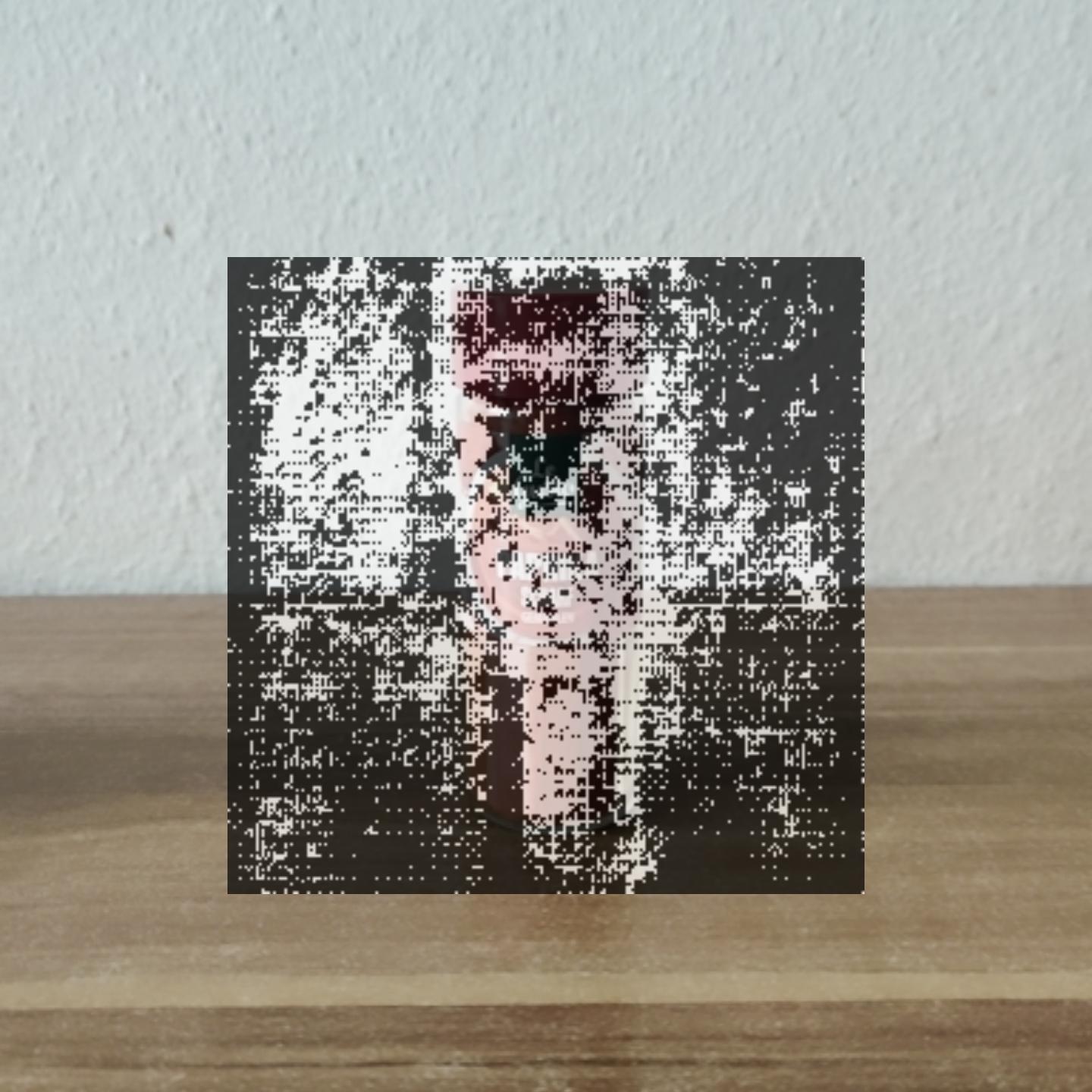}\\[4pt]
                \includegraphics[angle=90,trim=200 200 200 200, clip, width=0.9\linewidth]{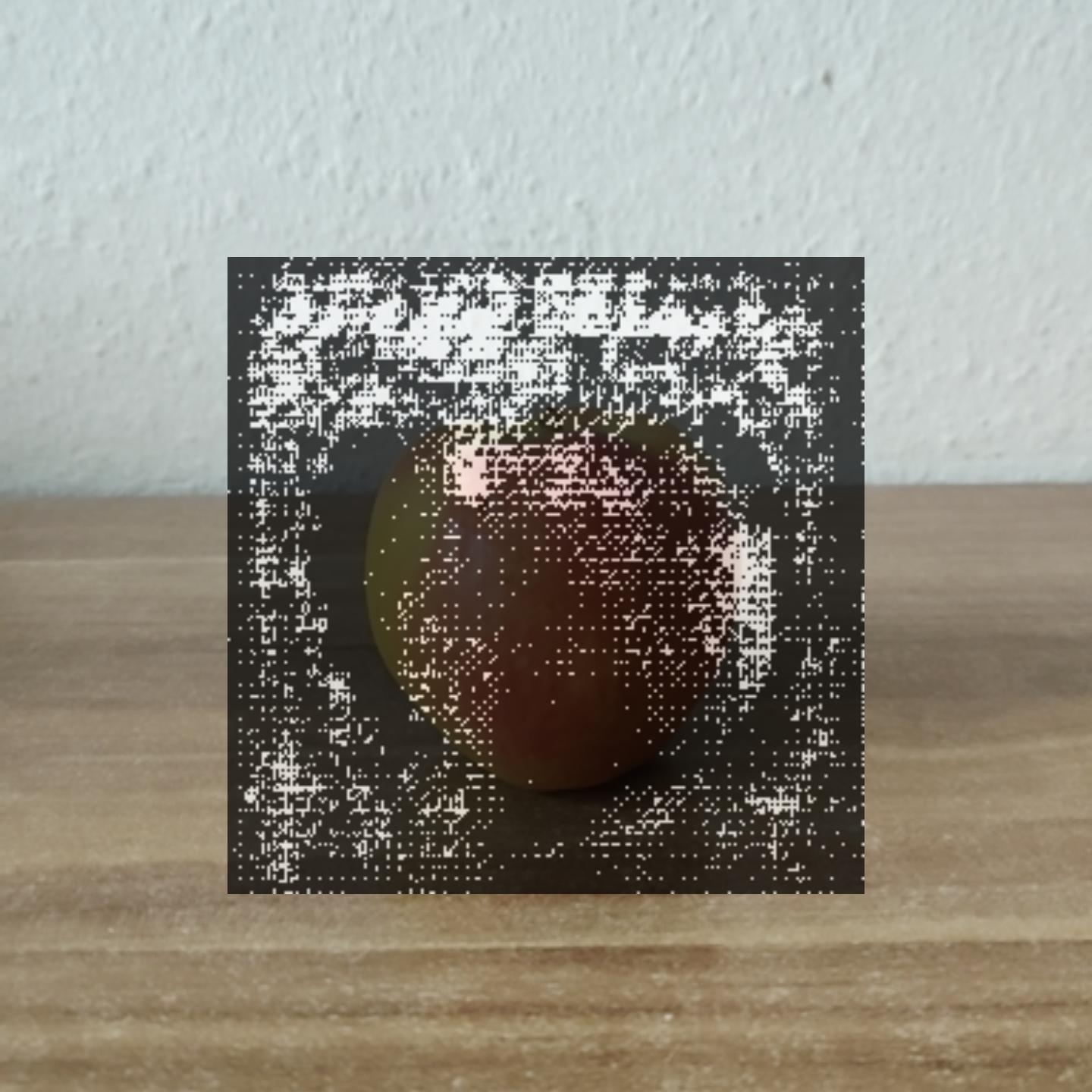}\\[4pt]
                \includegraphics[angle=90,trim=200 200 200 200, clip, width=0.9\linewidth]{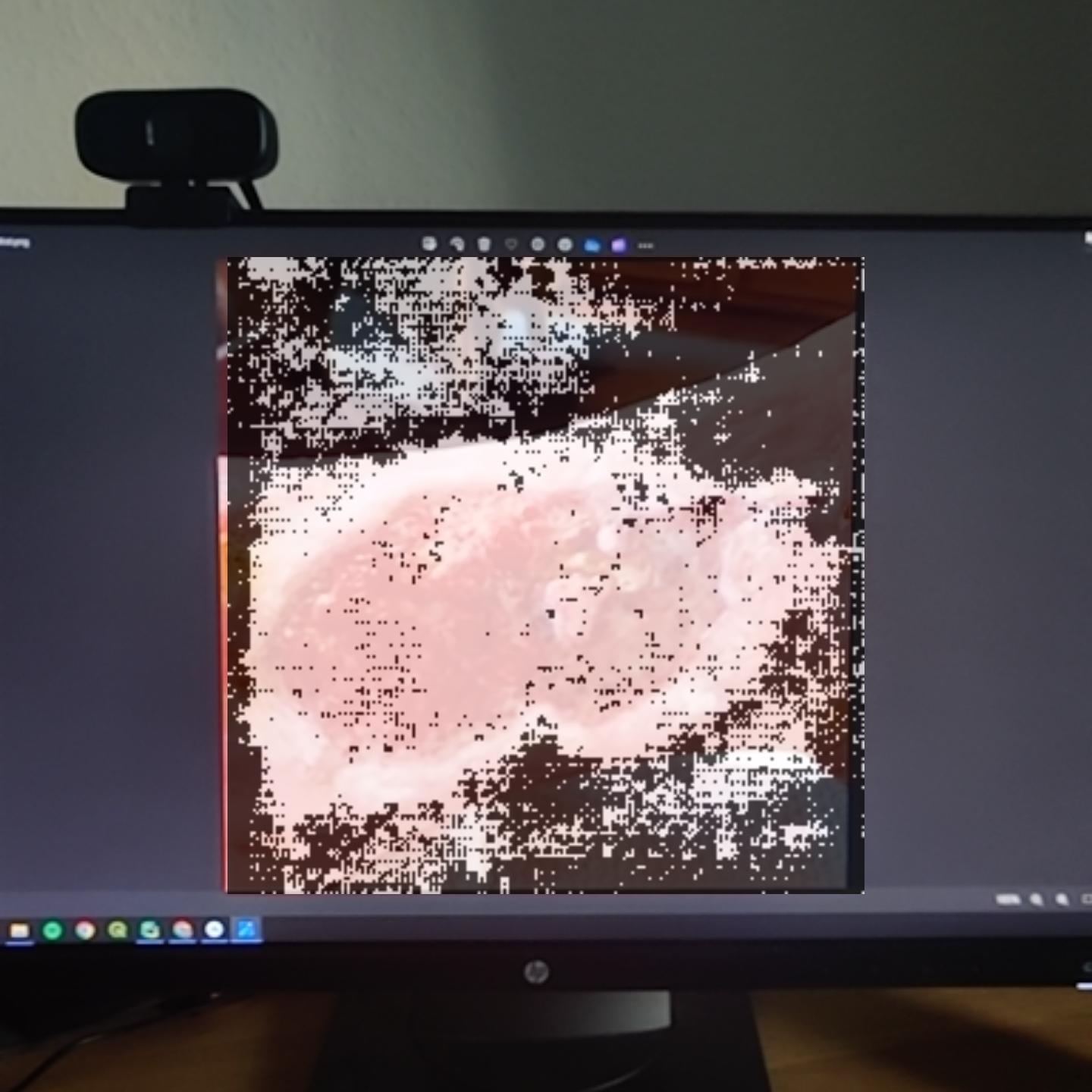}\\[4pt]
                \includegraphics[angle=90,trim=200 200 200 200, clip, width=0.9\linewidth]{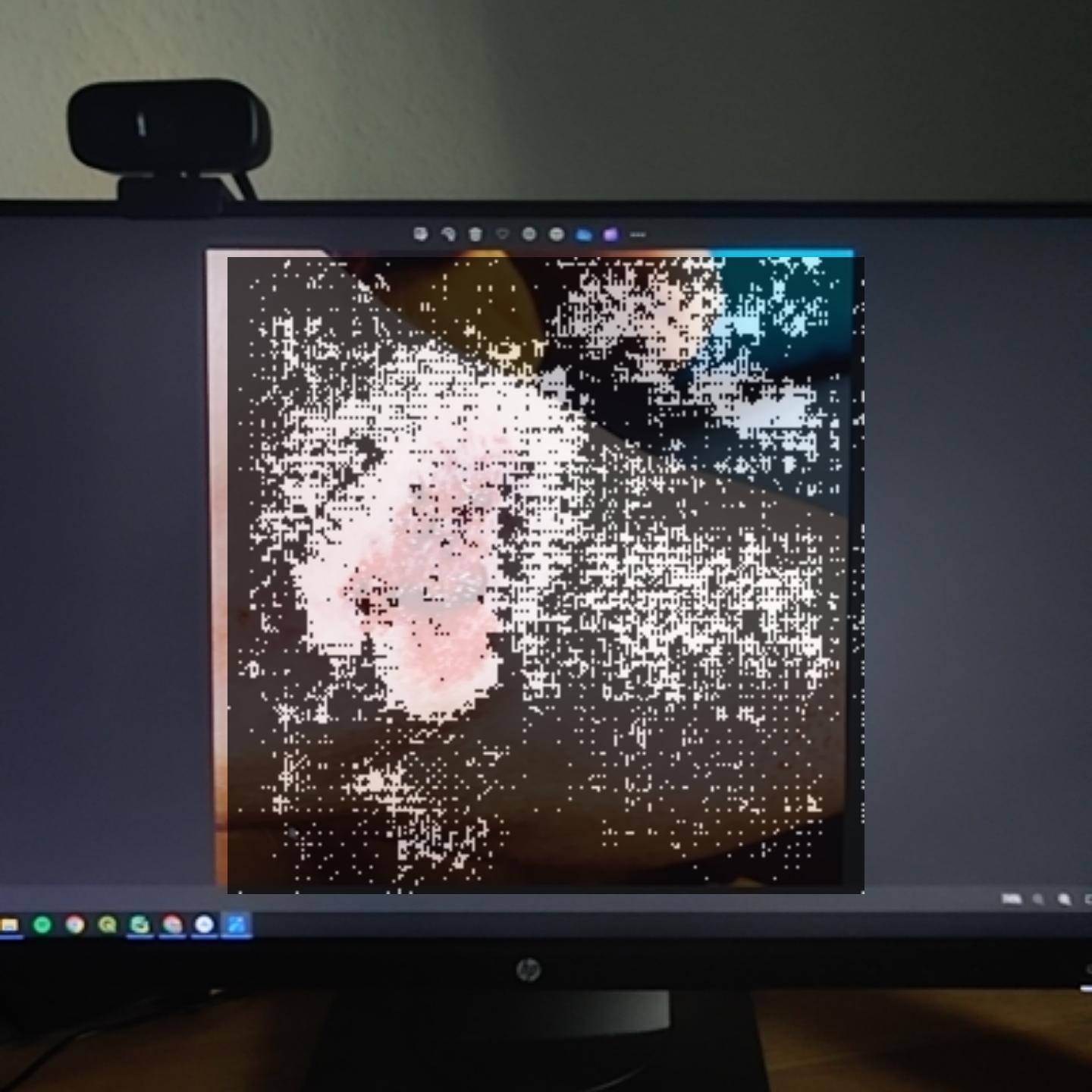}\\[4pt]
                \includegraphics[angle=90,trim=200 200 200 200, clip, width=0.9\linewidth]{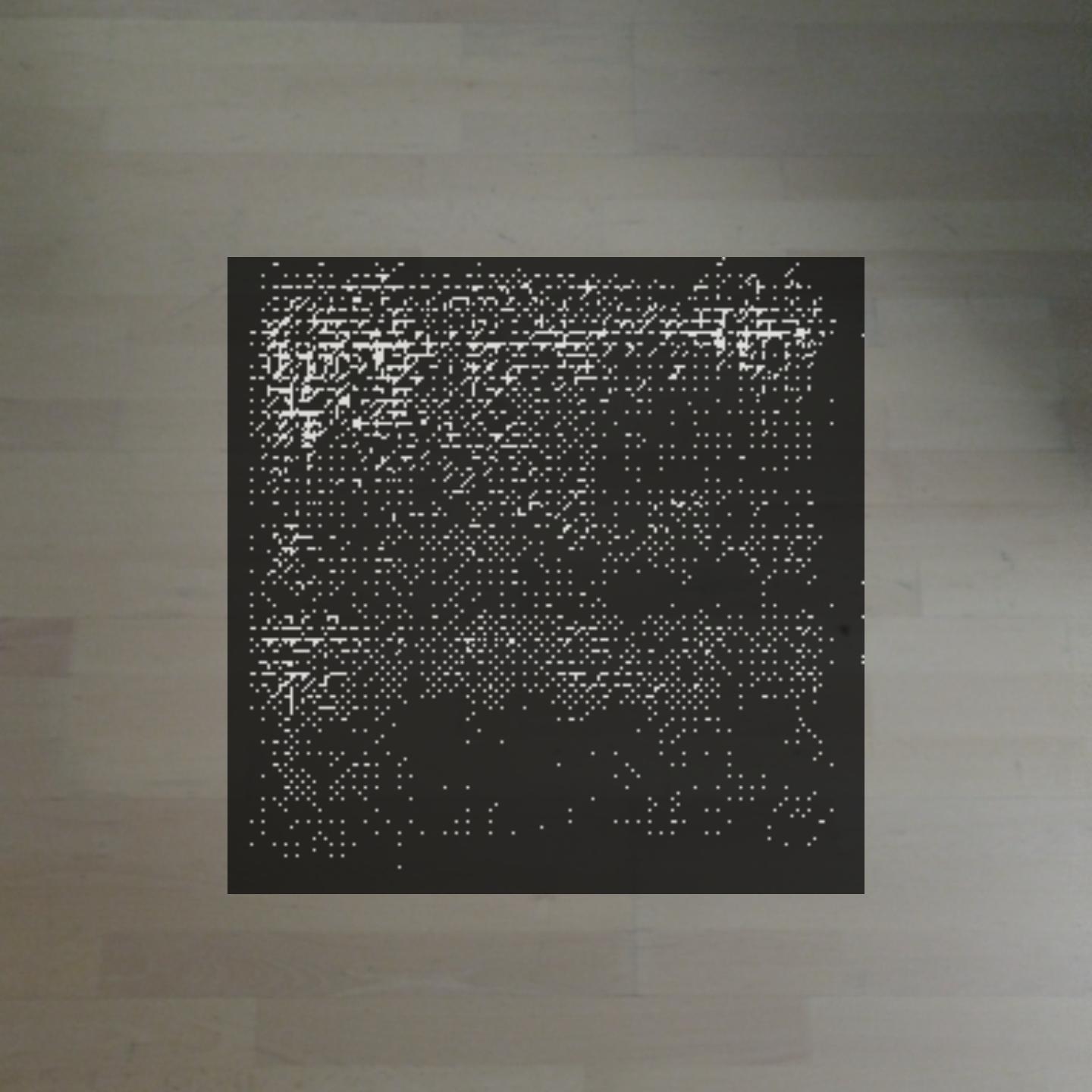}\\[4pt]
                \caption*{\scriptsize ENet}
            \end{subfigure}
            \begin{subfigure}{0.165\textwidth}
                \centering
                \includegraphics[angle=90,trim=200 200 200 200, clip, width=0.9\linewidth]{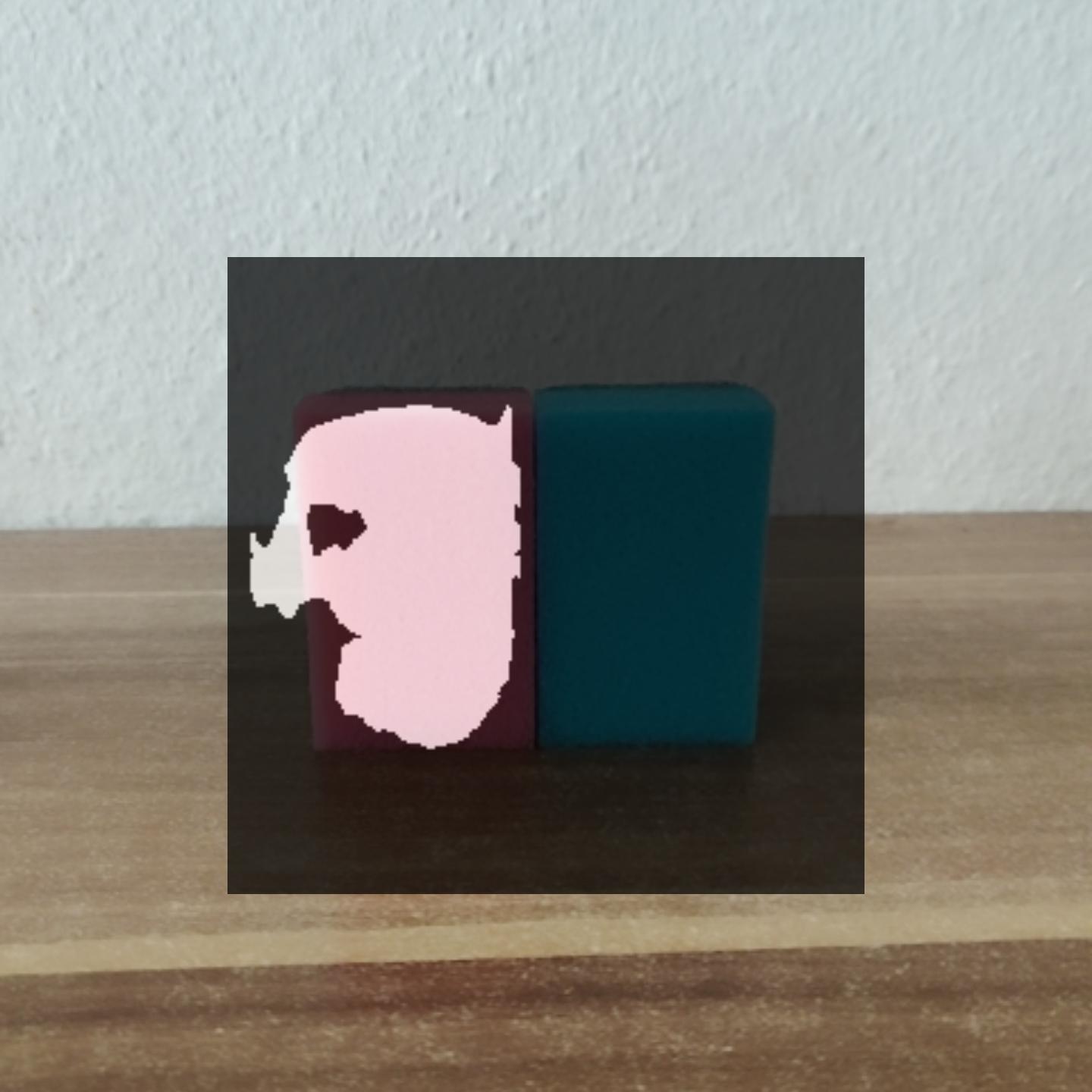}\\[4pt]
                \includegraphics[angle=90,trim=200 200 200 200, clip, width=0.9\linewidth]{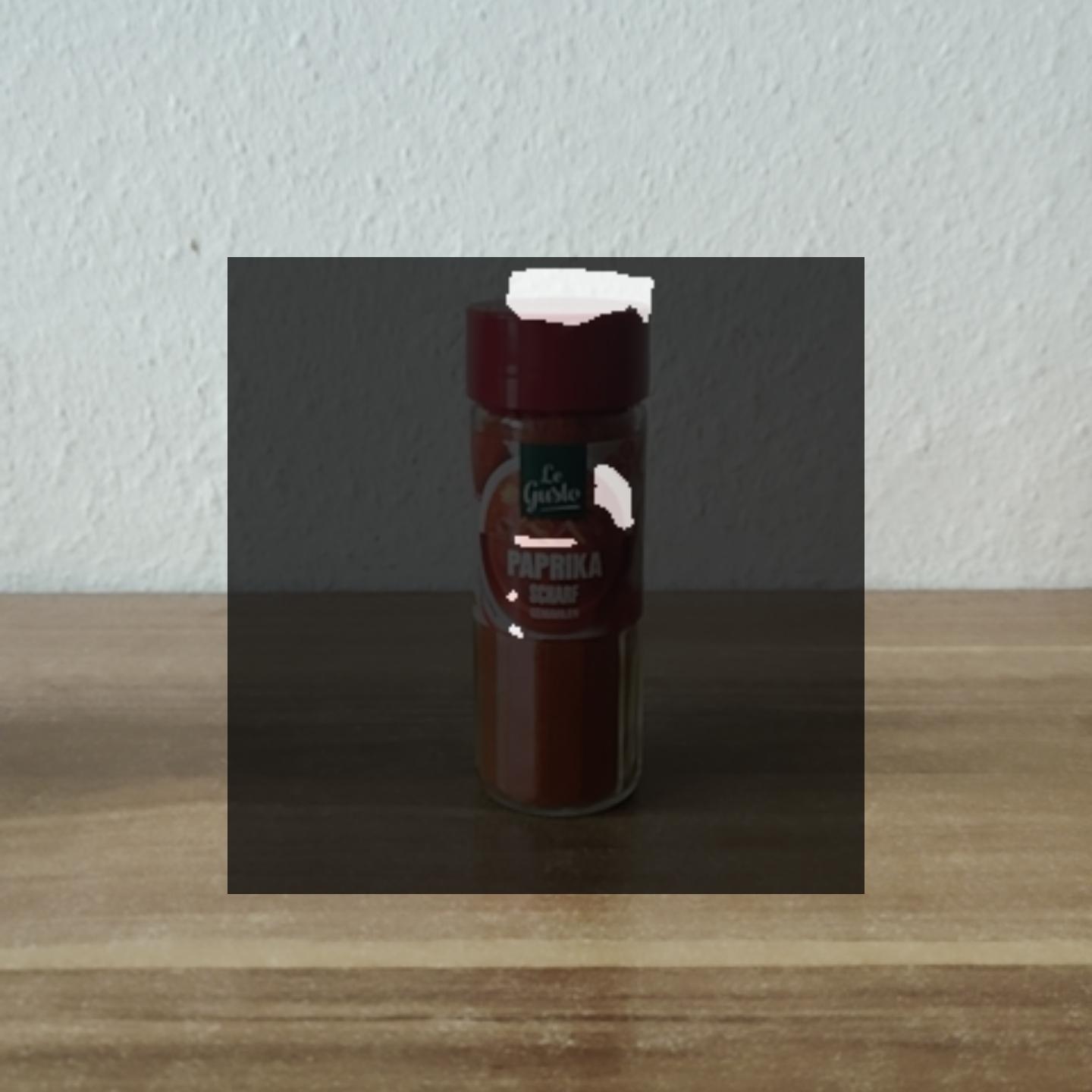}\\[4pt]
                \includegraphics[angle=90,trim=200 200 200 200, clip, width=0.9\linewidth]{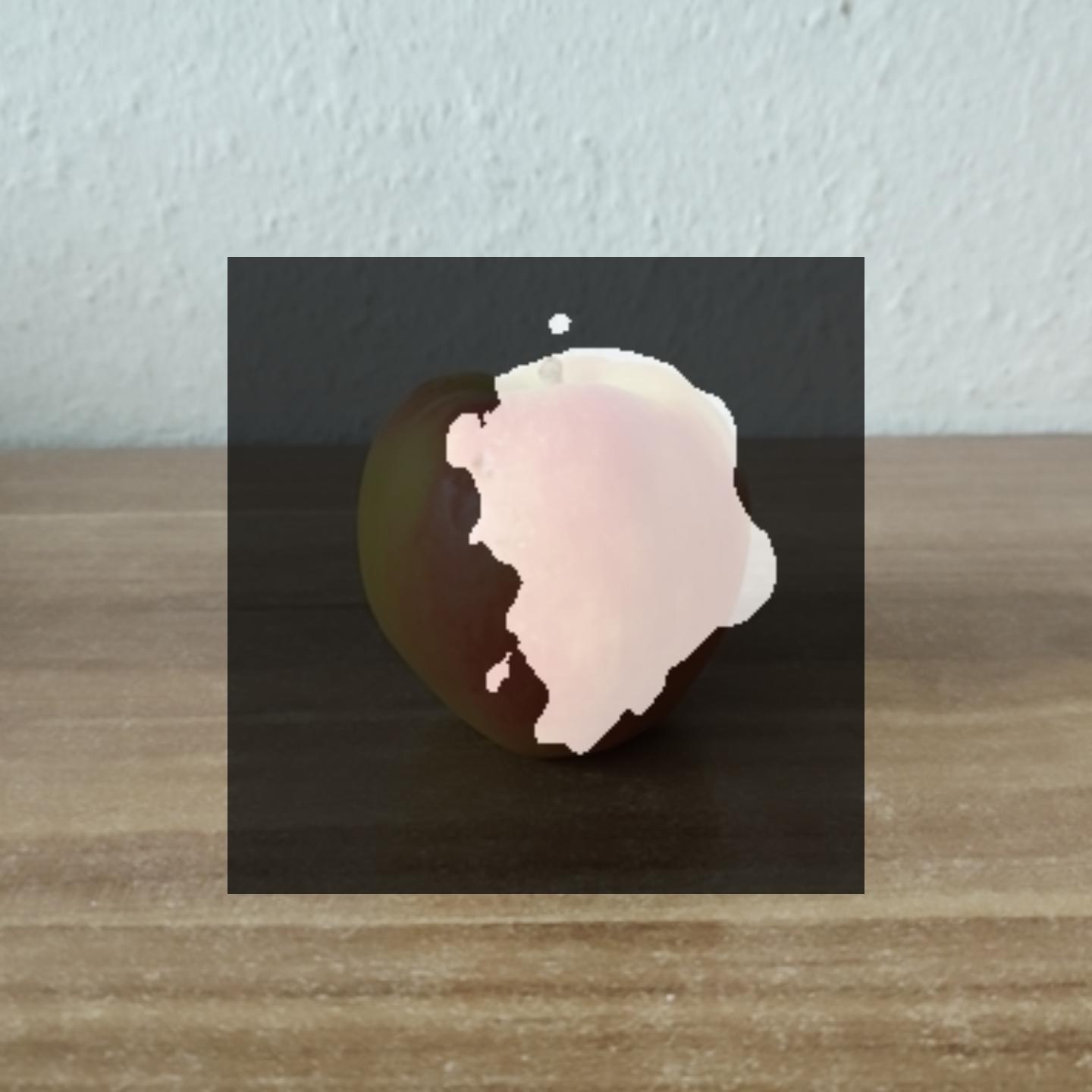}\\[4pt]
                \includegraphics[angle=90,trim=200 200 200 200, clip, width=0.9\linewidth]{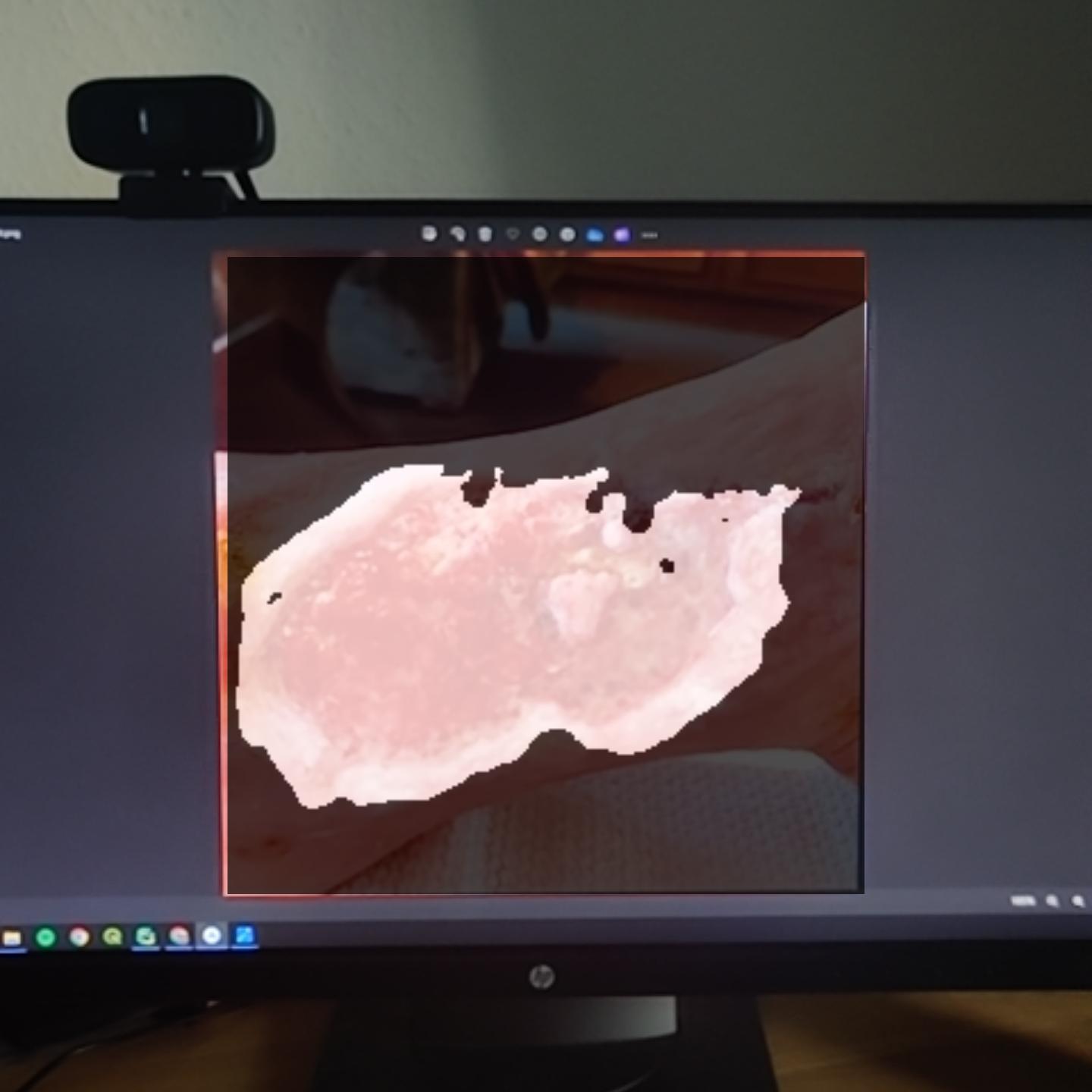}\\[4pt]
                \includegraphics[angle=90,trim=200 200 200 200, clip, width=0.9\linewidth]{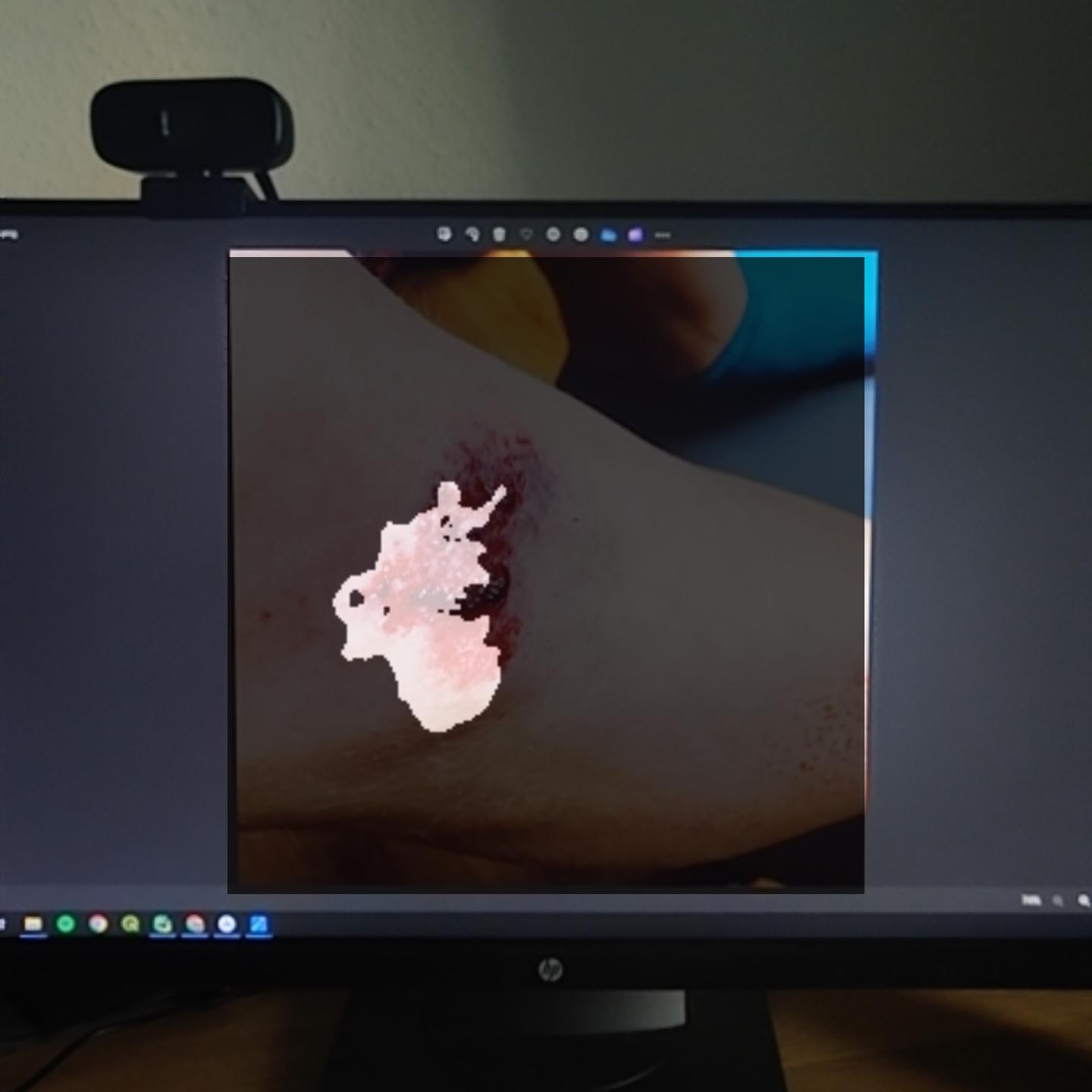}\\[4pt]
                \includegraphics[angle=90,trim=200 200 200 200, clip, width=0.9\linewidth]{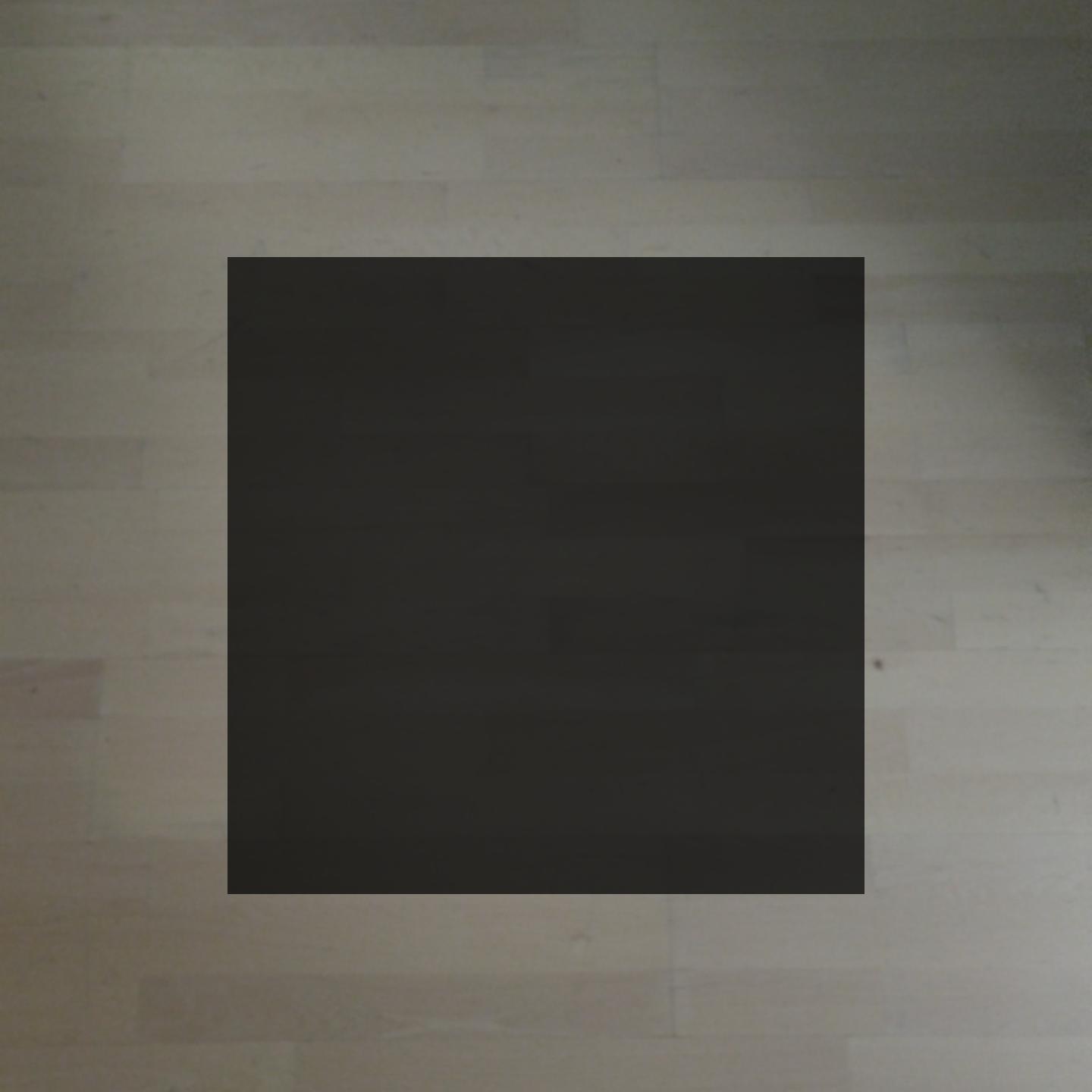}\\[4pt]
                \caption*{\scriptsize UNeXt-S}
            \end{subfigure}
        \end{adjustbox}
    \end{turn}
    \caption{Live segmentation of different model variants in our mobile app, arranged in descending order of parameter count.}
    \label{fig:eval:all_scenes}
\end{figure}

To assess the models' real-world capabilities, we developed a Flutter prototype. For seamless integration within the app, all PyTorch models except UNet were converted to TorchScript by tracing. The deployed networks performed live segmentation of camera streams, with center-cropped images to 224 × 224 pixels for efficient processing.
%Following preliminary trials, a threshold of 0.75 was established for the inference in favor of clearer segmentations and reduced jitter during prediction.
Following preliminary trials, a prediction threshold of 0.75 was established for the derivation of binary masks, with the objective of achieving more precise segmentations and reduced jitter during inference.
We rigorously tested the architectures on an \textit{OnePlus 7 Pro} Android phone from 2019. All models ran smoothly without stalling, demonstrating the feasibility of mobile segmentation using Flutter.
% Hinweis Konsti: Real time is schwierig hier allgemein, is ein delay von einer drittel Sekunde noch real time? Zwischendrin werden alle frames geskipped, wir nehmen nur einen neuen Frame zur Hand wenn die letzte prediction fertig ist
% Update Vanessa: Changed "real-time" to "live"
To enable visual assessments, we devised six distinct scenes and captured live segmentations via screenshots, utilizing the variants with pretraining. The scenes encompass: 
\begin{enumerate*}[label=(\roman*)]
\item a neutral environment devoid of objects
\item two diverse wound images not from the CFU dataset
\item two scenes featuring everyday objects with colors resembling wounds
\item a scene with two everyday objects, one of which has a wound-like color
% Good segmentation performance in real-world deployment to add value to daily medical practice/to create practical added value 
\end{enumerate*}.
%Example results are illustrated in Figure~\ref{fig:eval:all_scenes}, with more screenshots provided in Figures~\ref{fig:comparison:lightest_models} and~\ref{fig:comparison:unext_and_topformer}.
The resulting segmentation masks generated by the models are depicted in Figure~\ref{fig:eval:all_scenes}.
Visually, ENet exhibits the worst results among the tested models, with pixelated and holey masks as well as inaccuracies such as falsely identifying everyday objects, including the blue sponge. In contrast, the TopFormer and UNeXt variants show more promising results, identifying wounds with fewer imperfections, albeit with somewhat indistinct edges and contours. 
Notwithstanding the recognition difficulties encountered by the UNeXt models when presented with objects of wound-like colours, there is an overall improvement for both architectures compared to ENet. Overall, the TopFormer-based networks appear to be superior, as they proved quite effective at distinguishing between wounds and wound-coloured objects.  Notably, the smallest variant (TF-T) performs exceptionally well in our example scenes, while the CNN-based UNeXt-B, despite exhibiting promising performance in Table~\ref{tab:eval:comparison_performance}, displays significantly poorer performance.

\section{Conclusion and Future Work}
\label{sec:conclusion}
In conclusion, our preliminary investigations into lightweight neural networks demonstrate their potential for mobile wound segmentation. The analyzed architectures, especially TopFormer, yield promising results in real-world application scenarios, although some refinements are still necessary, such as clearer edge detection.  % or greater robustness against wound-colored objects.
Future work will include extending the comparison to further promising architectures such as MBSNet~\cite{jin2023novel}, %or hybrid CNN-ViTs like 
MobileFormer~\cite{chen2022mobile}, or Seaformer~\cite{wan2023seaformer}. Furthermore, retraining the networks on an expanded dataset with more images, including those featuring objects with wound-similar colors, is expected to enhance the robustness and generalization ability of the models, thereby improving the accessibility and effectiveness of future wound care.

%In conclusion, our preliminary investigations highlight the potential of lightweight deep neural networks for mobile wound segmentation, showcasing promising results in real-world deployment scenarios. Moving forward, further refinement of these models, particularly in addressing segmentation accuracy and robustness to diverse wound characteristics, is warranted. Additionally, exploration of novel architectures and optimization techniques tailored specifically for mobile wound segmentation holds promise for advancing the accessibility and efficacy of wound care in the future.
%Based on our evaluation results, it is evident that while ENet demonstrates suboptimal performance in wound segmentation, TopFormer and UNeXt variants show promise, albeit with some imperfections. Future work will entail extending the comparison to include hybrid CNN-ViTs like MBSNet and retraining the networks with an expanded dataset. This dataset will encompass more images, including those featuring everyday objects with colors resembling wounds, to enhance model robustness and generalization.

\begin{credits}
\subsubsection{\ackname} We thank Adrian Strutt for conducting preliminary experiments. % before the start of this project. 
\subsubsection{\discintname}
The authors have no competing interests to declare. % that are relevant to the content of this article.
\end{credits}
\clearpage

\begin{extended}
\appendix
\section{Appendix}

\begin{figure}[htb!]
    \captionsetup[subfigure]{labelformat=empty}
    \begin{subfigure}[t]{0.45\textwidth}
        \caption{\textbf{Examples from FuSeg}}
        \vspace{0.5cm}
        \begin{subfigure}{.5\textwidth}
            \centering
            \includegraphics[width=.95\linewidth]{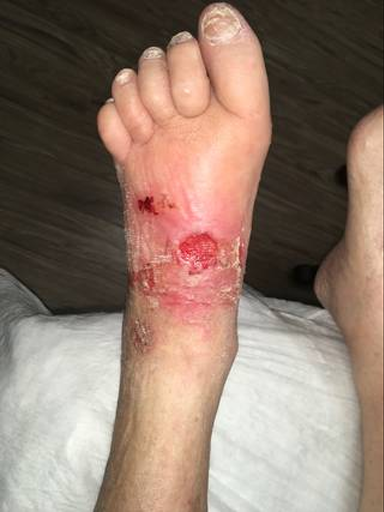}
            \caption{\footnotesize 0111.png}
        \end{subfigure}%
        \begin{subfigure}{.5\textwidth}
            \centering
            \includegraphics[width=.95\linewidth]{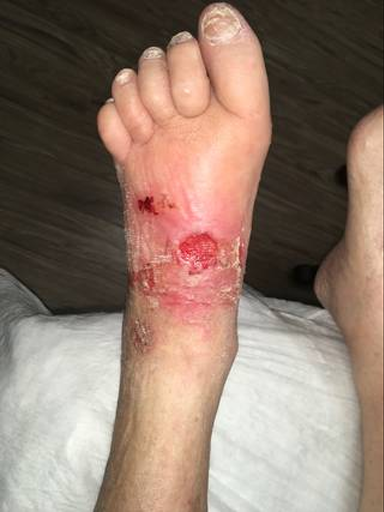}
            \caption{0386.png}
        \end{subfigure}    
    \end{subfigure}\hspace{\fill} % maximize horizontal separation
    \begin{subfigure}[t]{0.45\textwidth}
        % DFUC
        \caption{\textbf{Examples from DFUC}}
        \vspace{2cm}
        \begin{subfigure}{.5\textwidth}
            \centering
            \includegraphics[width=.95\linewidth]{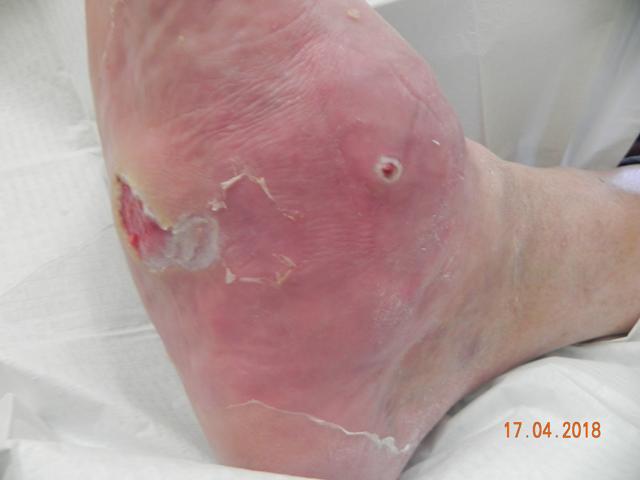}
            \caption{\footnotesize 100475.jpg}
        \end{subfigure}%
        \begin{subfigure}{0.5\textwidth}
            \centering
            \includegraphics[width=.95\linewidth]{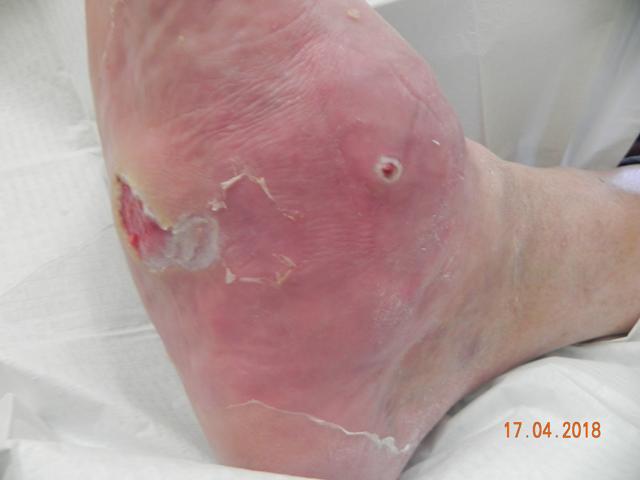}
            \caption{\footnotesize 100511.jpg}
        \end{subfigure}
    \end{subfigure}
    
    \bigskip % more vertical separation
    \begin{subfigure}[t]{0.45\textwidth}
        % FuSeg
        \begin{subfigure}{.5\textwidth}
            \centering
            \includegraphics[width=.95\linewidth]{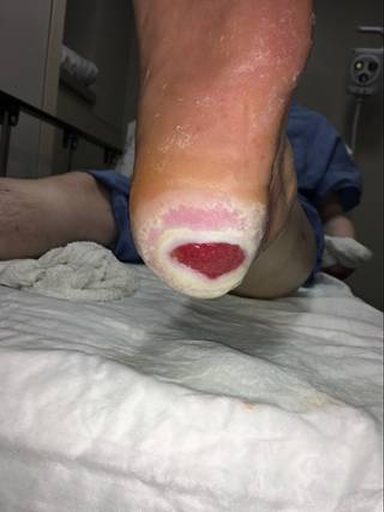}
            \caption{\footnotesize 0161.png}
        \end{subfigure}%
        \begin{subfigure}{.5\textwidth}
            \centering
            \includegraphics[width=.95\linewidth]{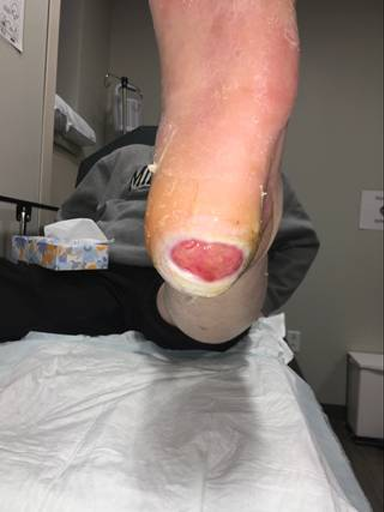}
            \caption{0163.png}
        \end{subfigure}
    \end{subfigure}\hspace{\fill} % maximize horizontal separation
    \begin{subfigure}[t]{0.45\textwidth}
        % DFUC
        \begin{subfigure}{.5\textwidth}
            \centering
            \includegraphics[width=.95\linewidth]{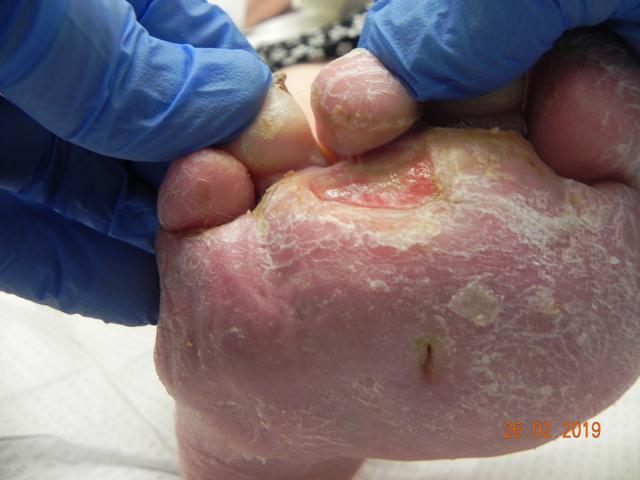}
            \caption{\footnotesize 101502.jpg}
        \end{subfigure}%
        \begin{subfigure}{.5\textwidth}
            \centering
            \includegraphics[width=.95\linewidth]{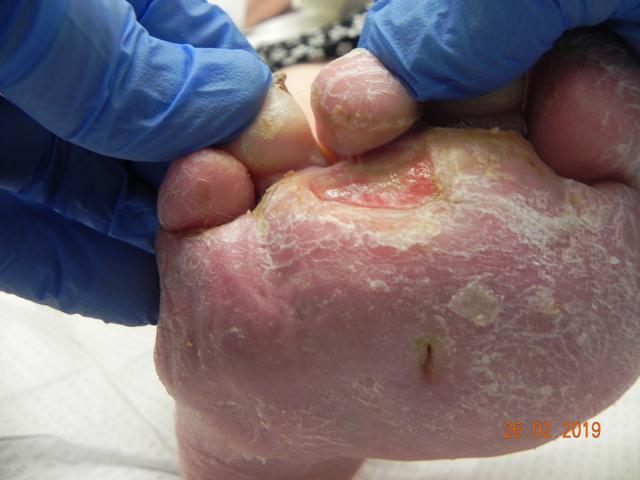}
            \caption{\footnotesize 101941.jpg}
        \end{subfigure}
    \end{subfigure}
    
    \bigskip % more vertical separation
    \begin{subfigure}[t]{0.45\textwidth}
        % FuSeg
        \begin{subfigure}{.5\textwidth}
            \centering
            \includegraphics[width=.95\linewidth]{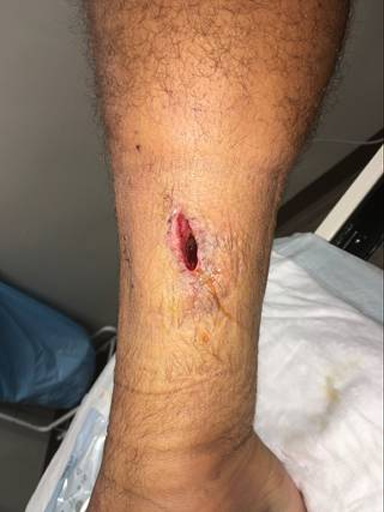}
            \caption{\footnotesize 0318.png}
        \end{subfigure}%
        \begin{subfigure}{.5\textwidth}
            \centering
            \includegraphics[width=.95\linewidth]{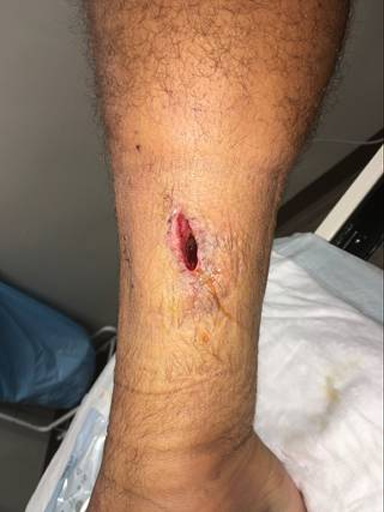}
            \caption{0415.png}
        \end{subfigure}
    \end{subfigure}\hspace{\fill} % maximize horizontal separation
    \begin{subfigure}[t]{0.45\textwidth}
        % DFUC
        \begin{subfigure}{.5\textwidth}
            \centering
            \includegraphics[width=.95\linewidth]{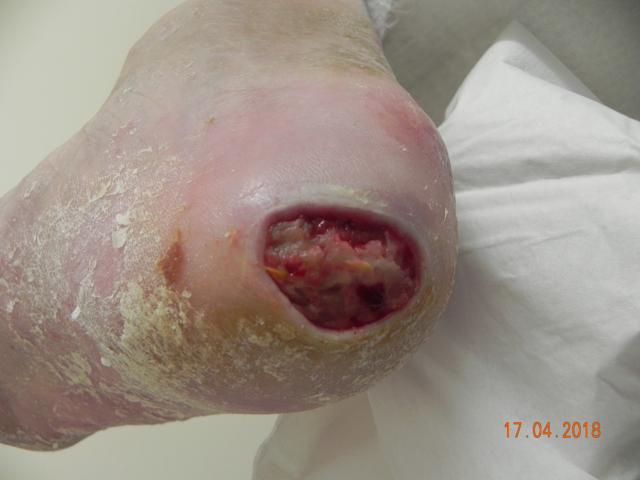}
            \caption{\footnotesize 100480.jpg}
        \end{subfigure}%
        \begin{subfigure}{.5\textwidth}
            \centering
            \includegraphics[width=.95\linewidth]{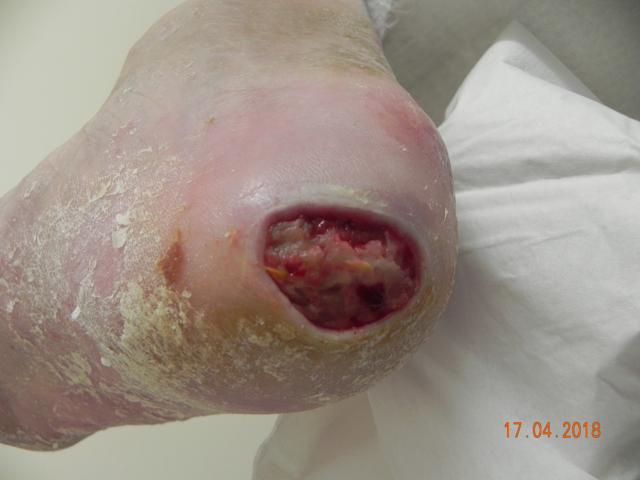}
            \caption{\footnotesize 100516.jpg}
        \end{subfigure}
    \end{subfigure}

    \caption{Examples of identical or highly similar image pairs from both datasets.}
    \label{fig:appendix:duplicates}
\end{figure}

\newpage
% ENet
\captionsetup[lstlisting]{hypcap=false}
\begin{multicols}{2}[
    \setlength\columnsep{2cm}
    \setlength{\columnseprule}{0.5pt}
    \renewcommand{\columnseprulecolor}{\color{black}}
    \captionof{lstlisting}{Example configuration for E-Net (pre-trained)}]
    \label{lst:appendix:enet}
\begin{lstlisting}[language=yaml, nolol=true]
architecture:
  arch_type: "ENetV2"
  model_name: "ENetV2-CFU-512-pretrained"
  in_size: 512
  use_pretrained: True
settings:
  max_epochs: 200
  batch_size: 4
  random_seed: 42
  full_determinism: True
  mixed_precision: True
  torch_compile: False
  num_workers: 4
  pin_memory: True
  checkpoints:
    save_mode: "best"
    metric: "iou"
    min_epoch: 30
  save_sample_segmentations: False
  transforms_backend: "torchvision"
optimizer:
  type: "AdamW"
  lr: 0.0001
loss:
  type: "default"
scheduler:
  type: "reduce_lr_on_plateau"
  metric: "dice"
  mode: "max"
  min_lr: 0.000001
  factor: 0.5
  verbose: True
early_stop: False
metrics:
  tracked:
    - "dice"
    - "iou"
    - "precision"
    - "recall"
  averaging_method: "micro"
  ignore_background_class: True
dataset:
  type: "CFU"
transforms:
  train:
    - to_tensor: True
    - pad_and_resize:
        size: 512
    - gaussian_blur:
        kernel_size: 25
        sigma: (0.001, 2.0)
    - threshold:
        threshold: 127
        pixel_min: 0
        pixel_max: 255
        apply_to_mask: True
        apply_to_image: False
    - color_jitter:
        brightness: 0.4
        contrast: 0.5
        saturation: 0.25
        hue: 0.01
    - normalize:
        mean: [ 0.5, 0.5, 0.5 ]
        std: [ 0.5, 0.5, 0.5 ]
    - random_horizontal_flip:
        p: 0.5
    - random_vertical_flip:
        p: 0.5
    - random_affine:
        degrees: (-180, 180)
        translate: (0.125, 0.125)
        scale: (0.5, 1.5)
        shear: 22.5
        fill: -1.0
        mask_fill: 0
  val:
    - to_tensor: True
    - pad_and_resize:
        size: 512
    - threshold:
        threshold: 127
        pixel_min: 0
        pixel_max: 255
        apply_to_mask: True
        apply_to_image: False
    - normalize:
        mean: [ 0.5, 0.5, 0.5 ]
        std: [ 0.5, 0.5, 0.5 ]

  test:
    - to_tensor: True
    - pad_and_resize:
        size: 512
    - threshold:
        threshold: 127
        pixel_min: 0
        pixel_max: 255
        apply_to_mask: True
        apply_to_image: False
    - normalize:
        mean: [ 0.5, 0.5, 0.5 ]
        std: [ 0.5, 0.5, 0.5 ]
\end{lstlisting}
\end{multicols}

\clearpage
\end{extended}

\clearpage
\bibliographystyle{splncs04}
\bibliography{paper}

\end{document}